\documentclass[sigconf,nonacm]{acmart}
\AtBeginDocument{%
  }

\copyrightyear{2026}
\acmYear{2026}
\acmConference[KDD '26]{Proceedings of the 32nd ACM SIGKDD Conference on Knowledge Discovery and Data Mining V.2}{August 09--13, 2026}{Jeju Island, Republic of Korea}
\acmBooktitle{Proceedings of the 32nd ACM SIGKDD Conference on Knowledge Discovery and Data Mining V.2 (KDD '26), August 09--13, 2026, Jeju Island, Republic of Korea}
\acmDOI{10.1145/3770855.3817484}
\acmISBN{979-8-4007-2259-2/2026/08}

\usepackage{multirow} 
\usepackage{subfigure}

\usepackage{enumitem}

\begin{document}

\title[Step-TP Dataset for Tensor Program Optimization]{Step-TP: A Grounded, Step-Level Dataset with Chain-of-Thought Reasoning for LLM-Guided Tensor Program Optimization}


\author{Mengfan Liu}
\affiliation{%
  \institution{The University of Hong Kong}
  \city{Hong Kong}
  \country{China}
}
\email{ml621@connect.hku.hk}

\author{Da Zheng}
\affiliation{%
  \institution{Ant Group}
  \city{Hangzhou}
  \country{China}
}
\email{zhengda.zheng@antgroup.com}

\author{Junwei Su}
\authornote{Corresponding authors.}
\affiliation{%
  \institution{University of Science and Technology of China}
  \city{Hefei}
  \country{China}
}
\email{junweisu.cs@gmail.com}

\author{Chuan Wu}
\authornotemark[1]
\affiliation{%
  \institution{The University of Hong Kong}
  \city{Hong Kong}
  \country{China}
}
\email{cwu@cs.hku.hk}

\renewcommand{\shortauthors}{Mengfan Liu, Da Zheng, Junwei Su, and Chuan Wu}

\ccsdesc[500]{Computing methodologies~Artificial intelligence}

\keywords{LLM-based Tensor Program Optimization, Tensor Program, Dataset, Chain-of-Thought Reasoning}

\newcommand{\mf}[1]{{\textcolor{blue}{mfliu: #1}}}

\begin{abstract}

Despite the strong reasoning capabilities of large language models (LLMs), optimizing the execution efficiency of tensor programs remains challenging due to the need for precise, composable transformation decisions. Recent LLM-guided approaches frame tensor program optimization as an iterative decision process, but existing datasets provide only end-to-end optimized program pairs using token-inefficient representations, lacking verifiable step-level supervision and interpretability. As a result, LLMs struggle to make reliable single-step decisions in large combinatorial optimization spaces. We introduce Step-TP, a post-training dataset for tensor program optimization that provides grounded, atomic, step-level supervision with structured chain-of-thought (CoT) reasoning. Step-TP forms a closed reasoning loop over intermediate program states, enabling reliable multi-step optimization rather than outcome imitation. Its design is guided by four principles: (i) a token-efficient, verifiable intermediate representation (IR) that deterministically lowers to TVM TIR; (ii) atomic and composable optimization strategies that decompose complex trajectories into interpretable single-step decisions; (iii) structured CoT supervision coupled with explicit IR-to-IR state transitions; and (iv) strategy filtering to balance coverage while preventing shortcut exploitation. The dataset and implementation are available at a GitHub link \url{https://github.com/LIUMENGFAN-gif/StepTP}.

\end{abstract}

\maketitle


\section{Introduction}

\textbf{Background and Motivation.} Efficient execution of deep neural networks on GPUs is fundamentally a problem of \emph{tensor program optimization}, in which high-level mathematical operators must be lowered into kernels that effectively exploit the GPU’s massive parallelism and hierarchical memory system~\citep{zheng2020ansor,abadi2016tensorflow,volkov2008benchmarking,vasilache2018tensor, wu2025mirage, zhong2025heta}. Although modern GPUs provide extraordinary peak throughput, realizing this performance in practice requires carefully coordinating computation, data movement, and parallel scheduling through techniques such as loop tiling, memory hierarchy selection, operator fusion, and thread–block binding. As model architectures become deeper, wider, and increasingly irregular, vanilla kernel implementations often lead to poor hardware utilization and excessive memory traffic, making performance highly sensitive to low-level code generation decisions~\citep{snider2023operator,tillet2019triton,li2022automatic,kim2020duplo,chen2018tvm,hu2024cdmpp}. This challenge is further amplified in contemporary workloads involving large language models (LLMs) and foundation models~\citep{liu2024deepseek, achiam2023gpt}, where even minor tensor-program-level inefficiencies accumulate into substantial increases in end-to-end latency, energy consumption, and deployment cost at scale. Consequently, efficient GPU execution has emerged as a central enabling factor for scalable, cost-effective, and sustainable deep learning, motivating the development of principled and automated tensor optimization techniques that can systematically reason about GPU execution behavior.

Motivated by these challenges, recent work has explored the use of LLMs as decision-makers in tensor program optimization, leveraging their capacity to reason over structured program representations and long-range dependencies in optimization sequences~\citep{zhai2024enabling, baronio2025kevin, tang2025reasoning, gong2025large, dong2025stark, li2025cuda, woo2025tritonrl, qi2026hetauto, liu2025optimizing}. Tensor optimization entails navigating a vast, discrete, and highly nonconvex search space composed of different-level decisions (e.g., graph, operator, memory, math levels), where effective strategies often depend jointly on local program structure and global execution context~\citep{jia2019taso,chen2018learning,baghdadi2019tiramisu, wu2025mirage, zheng2023einnet, wang2021pet, jia2019taso,zheng2023tileflow, zhu2022roller, fegade2024acrobat}. LLMs are particularly well suited to this setting because they naturally model optimization as a sequential decision process, integrating symbolic program information with high-level optimization intent to produce interpretable, step-wise transformations~\citep{tang2025reasoning,cummins2023large, gong2025large, dong2025stark, baronio2025kevin}. In contrast to black-box search methods or purely statistical cost models, LLM-based approaches offer the potential to incorporate domain knowledge, generalize across workloads and hardware backends, and reuse learned optimization patterns across programs. As tensor programs and accelerator architectures continue to grow in complexity, LLM-guided optimization provides a promising direction toward more flexible and data-efficient tensor optimization frameworks.

\noindent \textbf{Pressing Need for Post-Training Datasets.}
Despite recent progress, the practical effectiveness of LLM-based tensor optimization remains constrained without targeted post-training on domain-specific data~\citep{cummins2024meta,cummins2023large, woo2025tritonrl}. Although general-purpose LLMs exhibit strong reasoning and pattern recognition capabilities, they are not trained to manipulate low-level tensor programs under the strict correctness, hardware, and performance constraints required by real-world GPU execution.  In the absence of post-training data that explicitly encodes these requirements, LLMs tend to fall back on superficial pattern matching or outcome imitation, resulting in brittle behavior, poor generalization, and limited interpretability~\citep{chang2024language, zhai2024enabling, woo2025tritonrl, xu2026gac}. Meanwhile, reinforcement learning or search-based fine-tuning approaches are prohibitively expensive in this domain due to the high cost of compiling and benchmarking candidate programs~\citep{li2025cuda, baronio2025kevin, su2025cuda}. Consequently, a carefully constructed dataset for post-training—one that provides grounded intermediate program states, atomic optimization actions, and verifiable state transitions—becomes essential for enabling data-efficient learning, stable reasoning, and systematic evaluation. Such datasets are critical for transforming LLMs from coarse heuristic generators into reliable optimization agents capable of reasoning about tensor programs at scale.

\noindent \textbf{Desired Properties of Post-Training Datasets.}
To be effective, such post-training datasets must do more than simply collect optimized programs—they must be explicitly designed to support step-wise reasoning and iterative decision making in tensor optimization.
Tensor optimization is inherently an iterative decision process rather than a single-shot prediction task: it requires precise, step-wise reasoning over program representations, where each transformation must be valid, composable, and compatible with downstream steps~\citep{zheng2020ansor,chen2018learning, feng2023tensorir, jia2019optimizing, zhao2022apollo, zhao2021akg, wang2025tilelang, wang2022topoopt, wang2024ladder}. Accordingly, effective post-training for LLM-based tensor optimization demands datasets that (i) expose \emph{step-level supervision} instead of only final outcomes, (ii) provide \emph{faithfully grounded reasoning traces} aligned with executable program transformations, and (iii) cover a \emph{diverse set of optimization strategies} representative of real-world workloads.  In addition, as the inference and reasoning capacity of LLMs is constrained by context length, post-training datasets should (iv) encode optimization processes in a \emph{context-efficient representation} that supports effective reasoning within practical prompt-length limits.

\noindent \textbf{Limitations of Existing Datasets.}
Despite growing interest in this direction, existing datasets fail to jointly satisfy the desired requirements in several important respects. \textit{First}, most existing datasets rely primarily on outcome-only supervision~\citep{zhai2023tlp, zheng2021tenset, zhai2024enabling, merouani2025looperset, yangir}, providing only final high-performance tensor programs produced through complex compositions of multiple optimization strategies. Such supervision encourages LLMs to memorize surface patterns in optimized code rather than to internalize the underlying decision-making process, leading to weak reasoning ability, poor generalization to unseen programs, and limited capacity to explore novel strategy compositions.
\textit{Second}, existing datasets such as ConCuR~\citep{kong2025concur} adopt low-level CUDA or Python code as the primary representation space for optimization. While expressive, these representations are verbose and poorly suited for compactly encoding optimization intent and intermediate program states, resulting in excessively long descriptions that hinder effective reasoning within the limited context length of LLMs~\citep{li2024long,chang2024language,liu2024lost}.
\textit{Third}, existing datasets (e.g., LOOPerSet~\citep{merouani2025looperset}, IR-OptSet~\citep{yangir}, and ConCuR~\citep{kong2025concur}) exhibit limited strategy diversity, focusing mainly on easily modularized transformations such as loop tiling while largely omitting more sophisticated mathematical optimizations (e.g., online softmax~\citep{milakov2018online}). Because strategy diversity directly determines the effective optimization search space, this narrow coverage constrains an LLM’s ability to reason about, compose, and generalize high-performance solutions for real-world tensor programs~\citep{wu2025mirage,fang2021eto, jia2019taso, dao2022flashattention, dao2023flashattention, zheng2023tileflow}. Together, these limitations highlight a critical need for constructing datasets for LLM-based tensor program optimization that are \emph{representation-efficient}, support \emph{grounded, step-level supervision}, and enable \emph{reliable reasoning over a diverse set of optimization strategies}.

\noindent \textbf{Contributions.}
To address these limitations, we introduce \textit{Step-TP}, a post-training dataset for LLM-based tensor program optimization that provides grounded, atomic, step-level supervision with structured chain-of-thought (CoT) reasoning across diverse tensor-program–level optimization strategies. A comparison between \textit{Step-TP} and existing datasets is shown in Table~\ref{tab:dataset-comparison}. The main contributions of this paper are twofold:

\begin{enumerate}[leftmargin=*]
\item \textbf{Design of an Effective Intermediate Representation (IR).}
We propose LEIR, a \emph{verifiable and token-efficient intermediate representation} tailored for LLM-based tensor optimization. The LEIR supports compact and precise expression of intermediate program states, enables seamless application of atomic optimization strategies, and can be deterministically converted to TVM TIR, providing semantic grounding and correctness verification.
\item \textbf{Construction of a Step-Level Post-Training Dataset.}  
Building on the proposed IR, we construct \textit{Step-TP}, a post-training dataset for LLM-based tensor optimization that incorporates:  
(i) a systematic decomposition of complex optimization trajectories into \emph{atomic, composable strategies}, transforming a large and opaque search space into interpretable single-step decisions;  
(ii) \emph{structured CoT supervision} that couples strategy-level rationale with explicit IR-to-IR state-transition mappings; and  
(iii) a \emph{strategy filtering mechanism} based on preconditions, parameters, and synthesis depth to balance strategy distribution, ensure broad coverage, and prevent shortcut exploitation.
\end{enumerate}

\noindent
We further conduct an extensive empirical study of \emph{multi-step optimization via step-level guidance}. Our results show that \textit{Step-TP} enables effective step-level guidance, empowering diverse search paradigms to achieve strong performance with remarkable efficiency. Our results demonstrate that this guidance allows models to generate executable, grounded transformations across a diverse set of strategies and can navigate long-horizon optimization trajectories across various GPU architectures.


\begin{table*}[!t]
\vspace{-3mm}
\centering
\small
\begin{tabular}{lcccccc} 
\hline
Dataset & Task & Target Platform & Executable IR/Program& CoT & Strategy-Driven &Step-level Supervision\\\hline
LOOPerSet~\citep{merouani2025looperset}& Polyhedral compiler optimization& CPU/GPU&$\times$&$\times$ & $\times$& $\times$\\\hline
TenSet~\citep{zheng2021tenset} & Cost model &  CPU/GPU & $\bigcirc$ & $\times$ & $\times$& $\times$ \\\hline
Tlp~\citep{zhai2023tlp} & Cost model &  CPU/GPU & $\bigcirc$ & $\times$ & $\times$& $\times$ \\\hline
TpuGraphs~\citep{phothilimthana2023tpugraphs} & Cost model &  TPU & $\bigcirc$ & $\times$ & $\times$& $\times$ \\\hline
IR-OptSet~\citep{yangir} & Tensor program optimization & CPU & $\checkmark$ & $\times$ & $\times$& $\times$\\\hline
ConCur~\citep{kong2025concur} & Tensor program optimization & GPU & $\checkmark$ & $\checkmark$ & $\times$ & $\times$ \\\hline
Step-TP (Ours) & Tensor program optimization & GPU & $\checkmark$ & $\checkmark$ & $\checkmark$ & $\checkmark$ \\\hline
\end{tabular}
\caption{Representative Datasets for Tensor Programs. \textnormal{Our dataset Step-TP is the only post-training dataset for LLM-based tensor program optimization that provides grounded, atomic, step-level supervision with structured CoT across diverse optimization strategies.}
}
\label{tab:dataset-comparison}
\vspace{-6mm}
\end{table*}

\section{Design of Intermediate Representation (IR)}
This section presents the design of our IR for tensor-program-level transformations. We begin by examining why existing program representations are ill-suited for learning and reasoning about transformation logic, and distill a key structural insight from this analysis. This insight motivates a \emph{high-density loop–equation} representation that covers full tensor program optimization space, which we formalize as LEIR and illustrate through a concrete matrix multiplication case study.

\begin{figure}[!t]
 \centering
\subfigure[Core part of CUDA]{
   \includegraphics[width=0.45\textwidth]{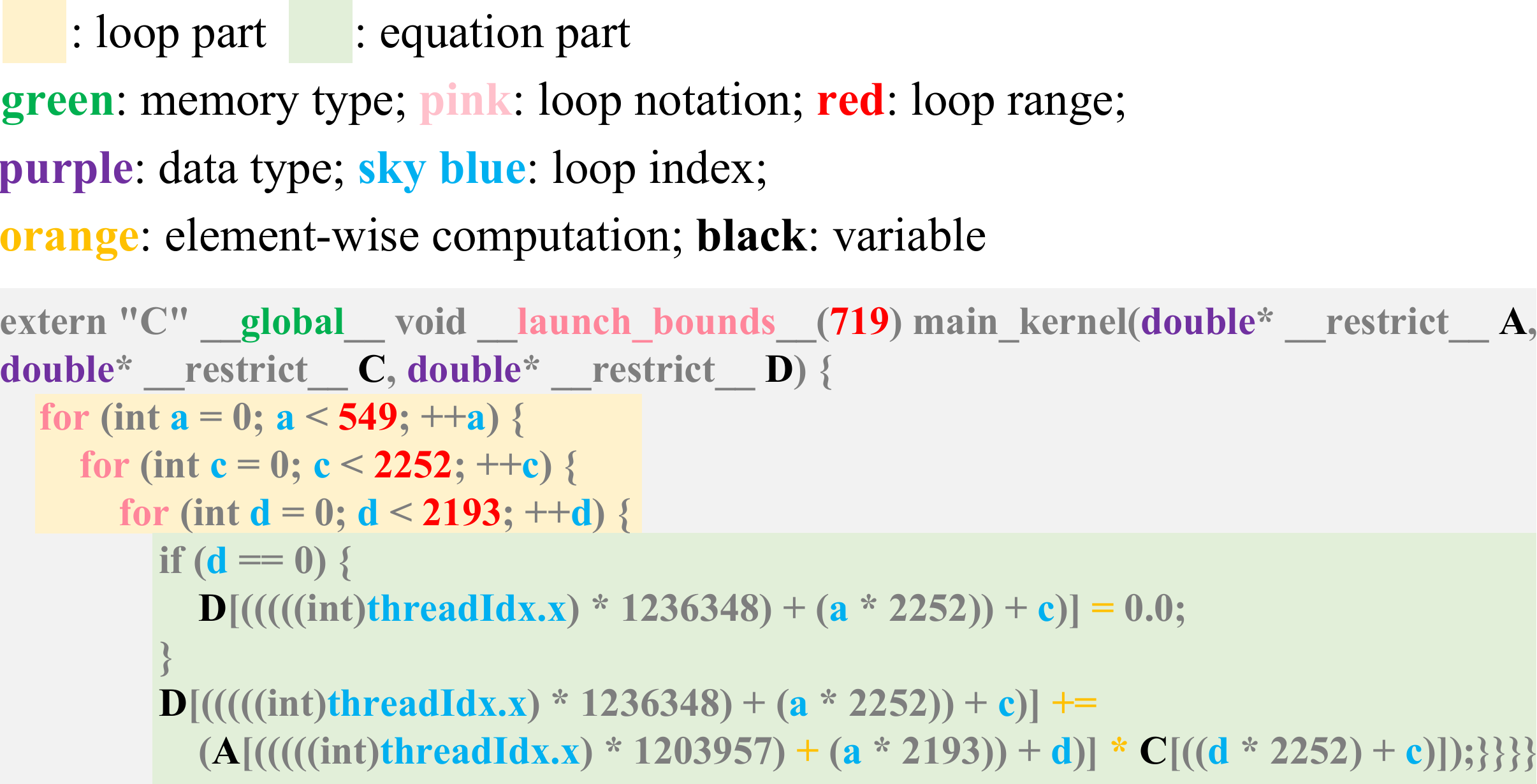}}\\
   \subfigure[TIR]{
   \includegraphics[width=0.45\textwidth]{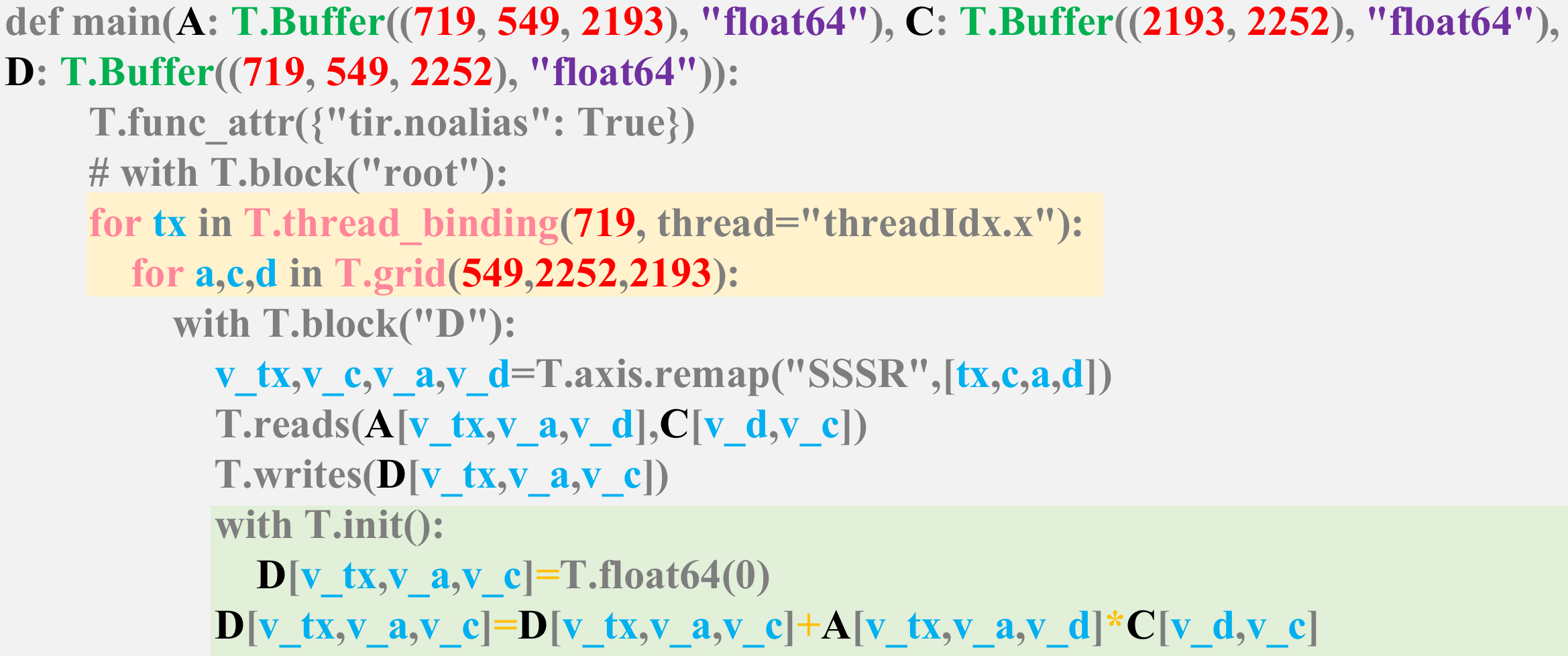}}\\
   \subfigure[Our LEIR]{
   \includegraphics[width=0.45\textwidth]{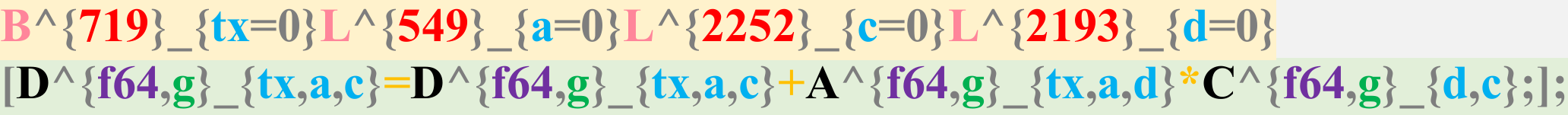}}
   \vspace{-4mm}
 \caption{Comparison of the same tensor program representation among CUDA, TVM TIR, and LEIR. \textnormal{As illustrated, our LEIR provides a more efficient representation for matrix multiplication ($D=D+A\times C$) and its associated loop structure.} 
 }
 \label{fig:IR-comparison}
 \vspace{-6mm}
\end{figure}

\subsection{Limitations of Existing IR.}
The reasoning capability of LLMs is constrained by finite context length, making representation efficiency a first-order concern. To enable effective modeling of program transformation logic, a tensor program representation should therefore be semantically dense and minimize entanglement with transformation-irrelevant details.

Mainstream representations such as CUDA and TVM TIR~\citep{feng2023tensorir} are executable and compiler-oriented, but they introduce substantial syntactic and structural noise—such as type annotations, control-flow scaffolding, and compiler-mandated boilerplate—that obscure the core transformation logic (e.g., loop restructuring and algebraic fusion). As a result, LLMs are overwhelmed with implementation artifacts that are orthogonal to tensor-program-level reasoning. We illustrate these limitations using a matrix multiplication example.

\noindent \textbf{CUDA.} As an explicit, hardware-oriented imperative model, CUDA prioritizes fine-grained control over GPU execution, thereby causing high-level transformation logic to be scattered across fragmented, implementation-specific constructs. As illustrated in Figure~\ref{fig:IR-comparison}(a), explicit type annotations in loop indices (e.g., \texttt{int a}, \texttt{((int)threadIdx.x)}) embed formatting details within the loops and index arithmetic. Meanwhile, manual initialization via control flow (e.g., \texttt{if (d == 0)}) structurally separates the initialization of a reduction from its accumulation update. Although executed within the same loop nest, this separation breaks the structural coherence of the reduction,  complicating the identification of the canonical matrix multiplication pattern ($e.g., D=D+A\times C$) as a unified transformation unit for learning-based models.
Furthermore, CUDA typically entangles the computation with micro-architectural designs such as memory bank-conflict avoidance. For instance, padding shared memory introduces intricate index offsets that obscure the logical iteration space, thereby introducing optimization concerns that are orthogonal to tensor-program-level transformations.

\noindent \textbf{TVM TIR.} Compared to CUDA, TIR offers a more structured representation aligned with tensor-program-level optimizations. However, as a compiler-oriented IR, TIR imposes a heavy burden of declarative boilerplate. As shown in Figure~\ref{fig:IR-comparison}(b), even a standard matrix multiplication is encased within extensive metadata (e.g., \texttt{T.reads}) and explicit axis-remapping mechanisms (e.g., \texttt{T.axis.remap}), which introduce substantial repetition without adding new semantic value to the algebraic computation.
Moreover, the core $D=D+A\times C$ logic is buried under multiple layers of syntactic scaffolding, such as the nested \texttt{T.block} and \texttt{T.init} scopes. These constructs, while essential for compiler correctness, create a high degree of structural depth that weakens the visibility of the fundamental transformation intent for learning-based models.

\emph{Therefore, neither CUDA nor TIR provides an efficient format for constructing transformation datasets. }

\subsection{Loop-Equation IR (LEIR).} 
To address these limitations, we examine the essential tensor-program-level structure common to both CUDA and TIR. After abstracting away low-level execution details and compiler boilerplate, both representations reduce to two irreducible semantic components:
\begin{enumerate}
\item a \emph{loop structure} defining the iteration space, and
\item an \emph{equation structure} specifying the algebraic computation at each iteration,
\end{enumerate}
where these two components still cover the full tensor-program-level optimization space detailed in Appendix~\ref{sec: appendix-strategy}, in contrast to prior abstractions (e.g., EINNET~\citep{zheng2023einnet}) that employ non-unified loop representations (hindering operator-level transformations like loop binding) and are restricted to summation-based computations.

This observation motivates an IR that preserves only these two components, yielding a representation that is both semantically dense and suitable for step-level learning.

\noindent \textbf{LEIR design principles.}
Building on this insight, we propose LEIR, a high-density representation that balances structural parsimony with the fidelity required to capture tensor-program-level transformations. Our design is guided by three core principles: 
\begin{enumerate}[leftmargin=*]
    \item \textbf{Irreducible semantic minimality.}
Unlike existing IRs that mandate extensive metadata for compiler analysis, LEIR distills the program representation into its minimal semantic components. It consolidates the fragmented constructs of CUDA and the multi-layered scaffolding of TIR into just two irreducible structures (i.e., nested loops and algebraic equations), thereby significantly enhancing the semantic density. This design ensures that the majority of tokens in the representation correspond directly to meaningful elements of the optimization space, rather than to syntactic overhead.
    \item \textbf{Explicit organization logic.} Explicit organization logic.
To maintain expressiveness, LEIR avoids the pitfall of excessive abstraction, such as representing programs solely as optimization parameters without the entire organizational logic of the computation. Instead, it explicitly preserves the structural hierarchy of execution. Specifically, the sequential order of loop descriptors and equations captures the execution flow. Meanwhile, the mapping of iteration spaces to logical execution levels (e.g., thread-block binding) is embedded within loop descriptors, and the assignment of data to memory hierarchies is encoded into tensor variables. This design ensures that the underlying computational pattern remains intact and reconstructible, allowing the learning-based models to reason about the spatial and temporal organization of the computation.
   \item \textbf{Parseable syntax.} LEIR adopts a LaTeX-based syntax to represent the tensor programs, which leverages the prior knowledge of LLMs to enable direct parsing and reasoning. Together, these principles enable LEIR to provide a concise yet expressive enough representation to capture tensor-program-level optimizations, while remaining interpretable by LLMs. 
   See Appendix~\ref{sec:appendix-IR} for the complete grammar definition.
\end{enumerate}

\noindent \textbf{Case Study.} As illustrated in Figure~\ref{fig:IR-comparison}(c), we exemplify the design of LEIR through a matrix multiplication case. A typical tensor program in LEIR consists of one or more expressions separated by semicolons, with each expression comprising a loop part (in yellow) and an equation part (in green). To ensure brevity, LEIR employs implicit initialization to maintain a concise algebraic flow.

\textit{Loop structure.} The loop structure is represented by a main symbol indicating the loop type, with superscripts and subscripts specifying the loop index and iteration range. LEIR supports various loop types: serial loops ($L$), parallel loops ($P$), vectorized loops ($V$), unrolled loops ($U$), and thread/block-binding loops ($B$). Notably, indices for binding loops are mapped to CUDA intrinsics: $\{bx, by, bz\}$ for \texttt{blockIdx} and $\{tx, ty, tz\}$ for \texttt{threadIdx}. In this case, $B^{719}_{tx=0}$ denotes the outermost loop bound to \texttt{threadIdx.x} with a range of 719, while $L^{549}_{a=0}$ represents a nested serial loop with a range of 549.

\textit{Equation structure.} The equation part specifies the computation performed under the given loop nest and consists of three elements: element-wise computation, tensor variables, and delimiters to define the computational scope and logical sequence.

(1) \emph{Element-wise computation.} We formalize the algebraic computations using a set of functional operators derived from TVM TIR, including standard arithmetic (e.g., $+$, $-$), transcendental functions (e.g., $\exp, \log$), and conditional intrinsics (e.g., $\text{if\_then\_else}$), all applied in a purely element-wise manner. The example illustrates the core matric multiplication operation: $D = D + A \times C$.

(2) \emph{Variable.} Each variable is defined by an identity symbol (e.g., $D$), with subscripts for indices and superscripts for metadata. The metadata includes the data type (e.g., $f64$ for float64) and memory hierarchy (e.g., $g$ for global, $s$ for shared, $l$ for local memory). For example, $D^{f64,g}_{tx,a,c}$ denotes a double-precision tensor variable stored in global memory.

(3) \emph{Delimiters.} Two types of delimiters are employed to organize the program structure. Specifically, the square brackets ($[$ and $]$) in the example bind the computation logic to the four-level loop nest. The internal semicolon ($;$) marks the completion of the expression, ensuring a clear logical sequence for operations.

\textit{Implicit Initialization.} To maintain an uninterrupted algebraic flow, LEIR employs implicit initialization for common reduction patterns. The identity element is automatically inferred from the operator: summations default to $0$, products to $1$, and extremes (max/min) to $\pm\infty$. Consequently, the accumulator $D^{f64,g}_{tx,a,c}$ is initialized to $0$ without requiring explicit code, maintaining a concise algebraic representation.

\section{Dataset Construction}

 \begin{figure*}[!t]
 \vspace{-4mm}
 \centering
   \includegraphics[width=0.8\textwidth]{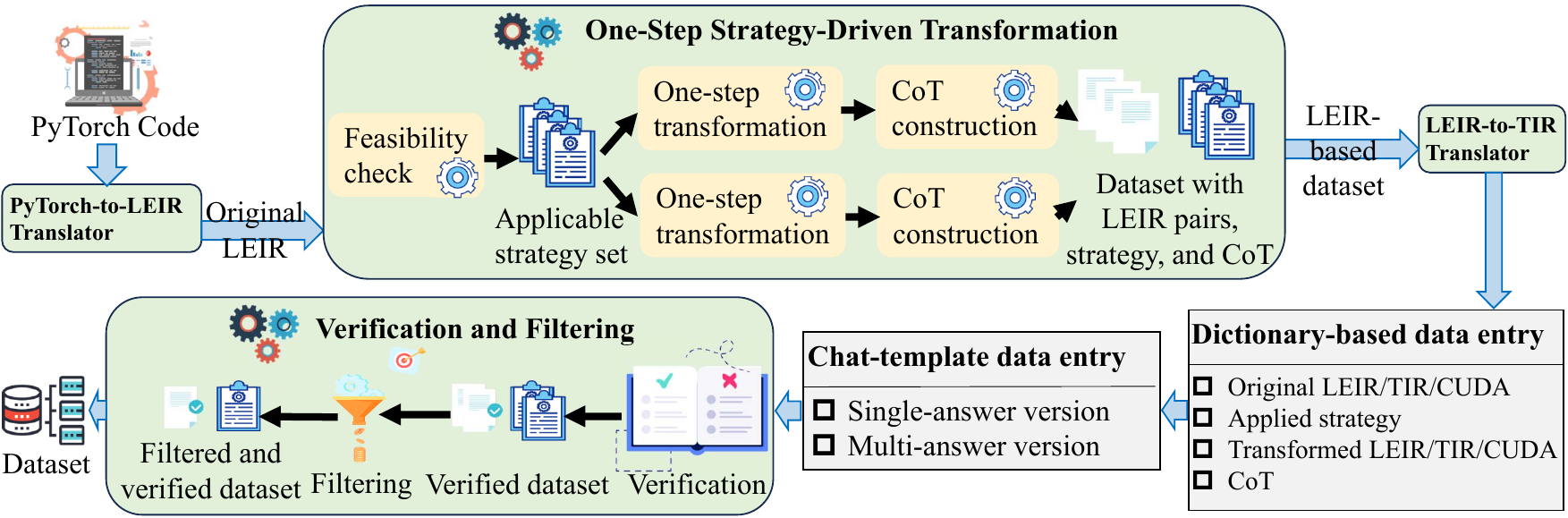}
   \vspace{-4mm}
 \caption{Pipeline of dataset construction.
 }
 \label{fig:pipeline}
 \vspace{-4mm}
\end{figure*}


To construct a high-quality dataset for grounded, step-level supervision of tensor program transformation with reliable reasoning, we design a multi-stage pipeline as shown in Figure~\ref{fig:pipeline}. Four stages are included: 
(i). The \textit{PyTorch-to-LEIR Translator} captures the computation in a PyTorch program 
and then converts it into our LEIR. 
(ii). The \textit{One-step Strategy-driven transformation} stage derives the applicable strategy set for each IR, applies strategies individually to perform single-step transformations, and generates a corresponding reasoning trace for each transformed LEIR to explain the optimization rationale. (iii). The transformed LEIR is then lowered to TVM TIR via an \textit{LEIR-to-TIR Translator}, enabling execution with a mature compiler backend. (iv). Finally, the \textit{Verification and Filtering} stage validates the correctness and semantic equivalence of transformed LEIRs, and applies a designed filtering mechanism to regulate the strategy distribution in the final dataset.
While the translators provide the necessary infrastructure, the Transformation and Filtering stages constitute the core mechanisms, ensuring the reliability of reasoning traces and the data quality.

This section is organized as follows: Sec.~\ref{sec:3-1} outlines the overarching dataset composition; Sec.~\ref{sec:3-2} details the transformation stage; Sec.~\ref{sec:format} describes the specific data formats utilized for archival and training; and Sec.~\ref{sec:filter} covers verification and filtering mechanisms.

\subsection{Dataset Composition}\label{sec:3-1}


The dataset comprises source PyTorch programs constructed through a hierarchical approach. We first establish two fundamental building blocks: (1). single-operator programs, mainly based on KernelBench~\citep{ouyang2502kernelbench} level-1 dataset, covering computational backbones (e.g., matrix multiplication), nonlinear activations, lightweight operations (e.g., transpose), normalization and pooling, and common loss functions. 
(2) popular architectures, such as Matrix Multiplication and Softmax pipelines and attention modules (e.g., multi-head attention, multi-group attention). 
Based on these building blocks, we further construct composite programs by sampling and assembling 2-5 components from the aforementioned categories, covering the majority of KernelBench~\citep{ouyang2502kernelbench} level-2 programs and additional randomly composed cases. To further enhance data diversity, we randomize the input and output shapes (e.g., with dimension sizes up to 16,384), and vary data types (e.g., float16, float32, and float64). By instantiating the 189 distinct program types with these varied configurations, we ultimately produce a diverse dataset comprising 6,335 unique PyTorch programs, providing comprehensive coverage of representative tensor computation workloads.

Based on the dataset composition, we employ a PyTorch-to-LEIR translator to convert these PyTorch programs into our LEIR. Since PyTorch operations abstract away low-level execution details (e.g., loop nesting), we align the underlying program structures of our IR with the corresponding vanilla TVM TIR implementations, ensuring correctness, semantic equivalence, and executability.

\subsection{One-step Strategy-driven Transformation}\label{sec:3-2}

This subsection details the one-step strategy-driven transformation stage, which shifts from traditional end-to-end mapping to step-level supervision with reliable reasoning. Two benefits are included: (i) step-level supervision guides the LLM through individual transformation, reducing learning difficulty and enabling generalization to different combinations of optimizations; (ii) the strategy-driven design ensures the independence and composability of each transformation strategy, avoiding the lack of interpretability in end-to-end optimizations. 
To implement this stage, we employ \textit{a three-phase transformation process}: a feasibility check of strategy preconditions, a one-step transformation to generate target LEIR, and a CoT construction to trace the transformation logic.

\noindent \textbf{Feasibility Check.} To ensure the validity of the generated candidate, each tensor program undergoes a feasibility check to identify applicable transformation strategies. 
We define nine essential preconditions, such as pattern-matching checks (e.g., identifying softmax for online softmax), with all preconditions and their corresponding strategy mappings in Appendix~\ref{sec: appendix-strategy}. By checking these preconditions, we establish a set of feasible strategies for each LEIR.

\noindent \textbf{One-step Transformation.} Based on the feasible strategy set for each original LEIR, we generate one-step transformed LEIRs via a decompose-modify-reassemble workflow to control the scope of modifications and improve reproducibility.
The decomposition follows the structure of our LEIR, enabling strategies at different levels (i.e., graph, operator, memory, and mathematical) to modify only the relevant components before reassembling them into a transformed LEIR. 
For example, the log simplification strategy works solely on the equations, while the loop reorder strategy acts only on the loops. When a strategy admits multiple valid outcomes (e.g., different loop split factors), we randomly sample one variant to enhance the diversity of transformations covered by each strategy.

\noindent \textbf{CoT Construction.} After applying the strategy, 
a corresponding reasoning trace is synthesized to formalize the underlying transformation logic. As shown in Fig.~\ref{fig:data-example}(a), each reasoning trace comprises two components: (i) a brief description of the applied strategy to provide a high-level semantic anchor for the transformation, and (ii) an instance-specific explanation to delineate the targeted expressions and components (e.g., loop or equation segments) and document their reassembly into the resulting modified expressions.

This structured CoT design yields several benefits for learning and interpretability. First, it improves the LLM’s understanding of program representations and enhances generalization to unseen LEIRs by explicitly constructing the reasoning with the underlying LEIR structure. 
Second, it facilitates the activation of tensor program optimization knowledge acquired during pretraining by presenting transformations in a strategy-centric and interpretable form. 
Third, the reasoning traces are directly derived from the actual transformation process rather than post-hoc rationales, and can be explicitly mapped to different-level modifications, ensuring full transparency and traceability.

After transformation, the LEIR-to-TIR translator maps both the original and transformed LEIRs into TVM TIRs, leveraging the TVM backend to generate executable CUDA kernels for end-to-end performance evaluation.

\subsection{Dataset Format}\label{sec:format}

 \begin{figure}[!t]
 \centering
\subfigure[Single-answer]{
   \includegraphics[width=0.47\textwidth]{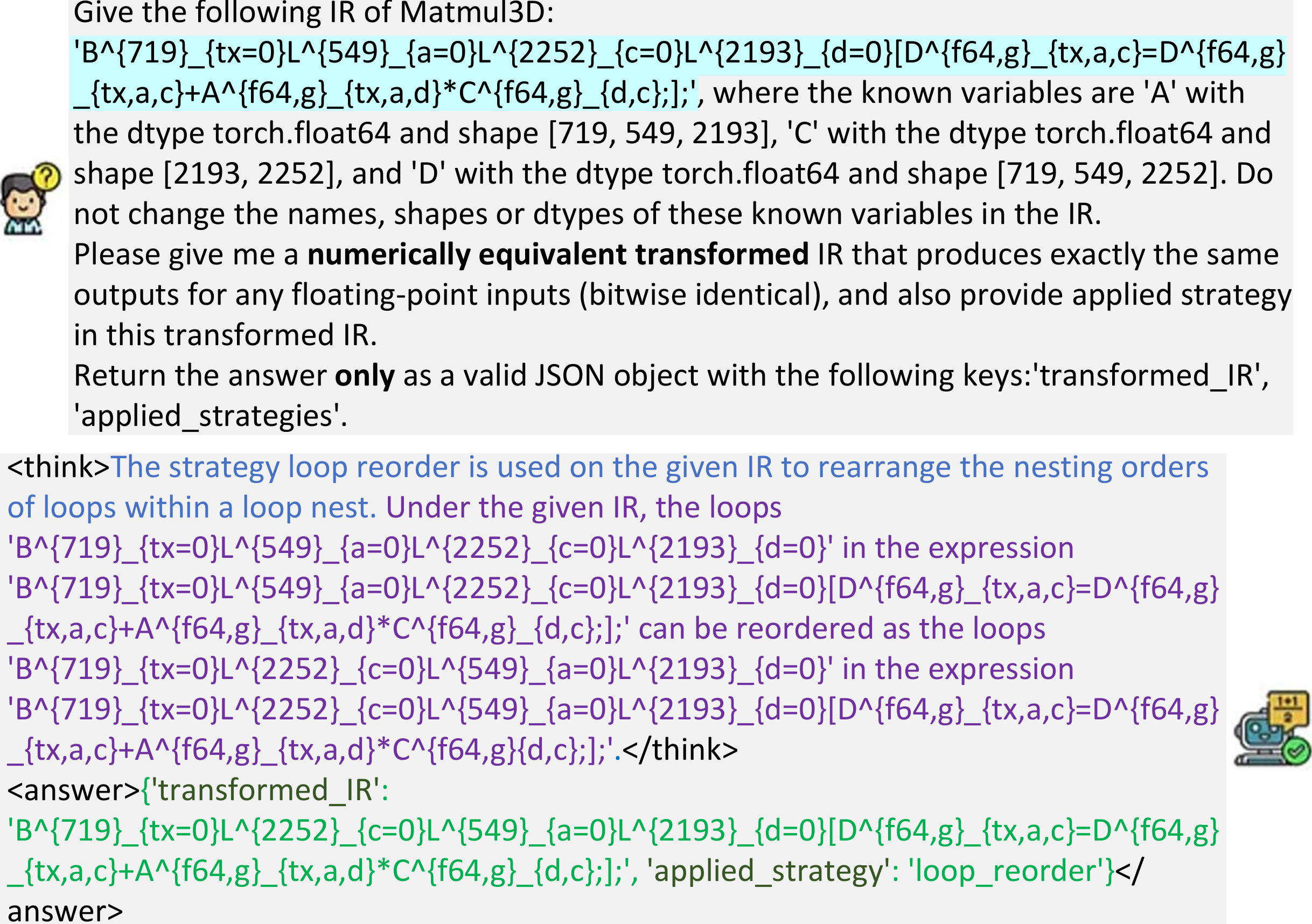}}\\
   \subfigure[Multiple-answer]{
   \includegraphics[width=0.47\textwidth]{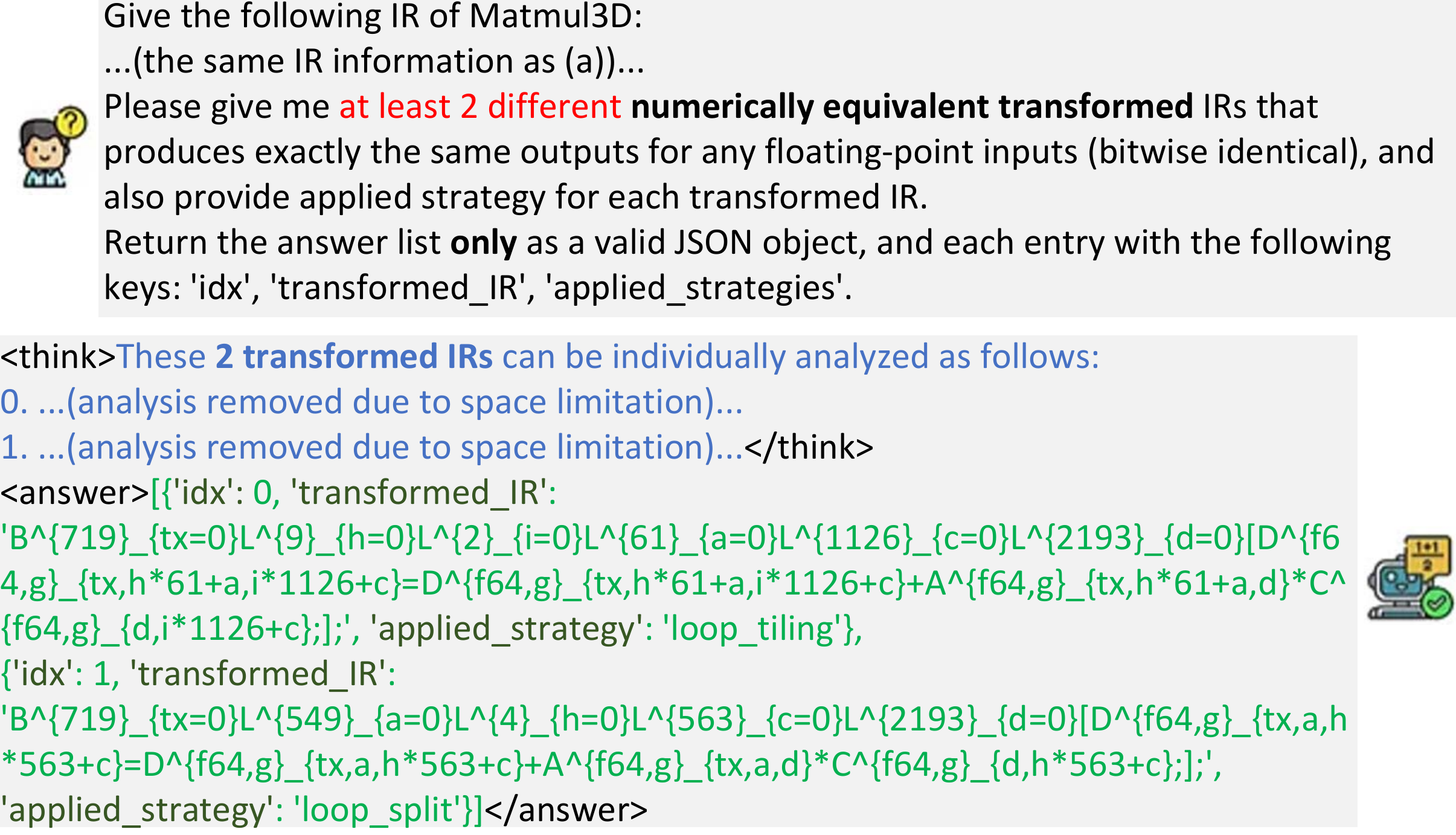}}
   \vspace{-6mm}
 \caption{Single-answer and multi-answer examples for Matmul3D in the chat-template dataset.
 }
 \label{fig:data-example}
 \vspace{-7mm}
\end{figure}

After the IR-to-TIR translation, we organize our data into two distinct formats: a comprehensive \textit{dictionary-based} repository for archival purposes and a \textit{chat-template} dataset for model training.

\noindent \textbf{Dictionary-based Dataset.} In this format, each data entry contains the original and transformed versions of the LEIR, TIR, and CUDA code, along with the applied strategy and the corresponding CoT. While this dictionary supports various program representations for future research, this paper mainly focuses on our LEIR.

\noindent \textbf{Chat-template Dataset.} Based on the dictionary repository, we construct a chat-template dataset, containing two specialized variants for training: (1).\textit{Single-answer format} pairs an original LEIR with one specific transformation; (2). \textit{Multi-answer format} incorporates a randomized subset of multiple transformations for an original LEIR. This multi-answer design not only encourages diverse optimization reasoning but also accommodates various multi-step optimization scenarios, such as providing multiple candidates for node selection in beam search.

As illustrated in Figure~\ref{fig:data-example}, both variants follow a standardized prompt-label architecture. The prompt integrates the original LEIR, essential metadata (e.g., program name, input/output shapes and data types), task specifications, and format requirements. The label comprises the CoT trace and the final answer. Specifically, in the multiple-answer variant, the CoT summarizes the number of transformed IRs and provides a numbered reasoning trace for each.

\subsection{Verification and Filtering Mechanism}\label{sec:filter}

To guarantee dataset correctness and maintain a balanced strategy distribution, we implement a two-fold pipeline consisting of empirical verification and strategy-difficulty-aware filtering.

\noindent \textbf{Verification.} To ensure the reliability of our dataset, we subject all original and transformed LEIRs to a rigorous verification process. For each program, we execute three independent trials using randomized input tensors and compare the outputs against the baseline. Only programs that exhibit consistent numerical equivalence across all trials are retained for the final dataset.

\noindent \textbf{Filtering.}
While the verification ensures functional correctness, it does not guarantee a high-quality distribution of transformation patterns. 
Without explicit control over strategy distribution, LLMs tend to disproportionately favor simplistic transformations. 
This bias stems from an asymmetry between simple and complex strategies. Simple strategies (e.g., log simplification) are frequently encountered during pre-training and can be activated with minimal supervision, resulting in lower predictive entropy and higher generation confidence. In contrast, complex strategies (e.g., loop split with index remapping and range recalculation) exhibit higher 
variability and structural diversity, which increases predictive uncertainty and causes the model to systematically avoid them.
To mitigate this bias and encourage the LLMs to master sophisticated reasoning, we implement a difficulty-aware rebalancing approach.

\textit{Strategy difficulty formulation.} To operationalize this rebalancing, we formalize the difficulty of each strategy along three dimensions:
\begin{itemize}
    \item Preconditions ($K$): the number of essential constraints identified during the feasibility check in Sec.~\ref{sec:3-2};
    \item Parameter modification ($P$): the number of existing components modified in the original IR, covering six aspects (i.e., expressions, loop axes, range adjustments, equations, variables, and index calculation segments);
    \item Synthesis depth ($S$): the number of unique categories of newly introduced elements (including new variables, index calculations, and expressions). Each category contributes a fixed value of 1 to the depth score.
\end{itemize}

These three dimensions reflect an ascending complexity, ranging from static constraint checking ($K$) and structural modification ($P$) to the generative synthesis of new logic ($S$). Therefore, the final difficulty score is formulated as:
\begin{equation}
    \text{difficulty score}=0.1K+0.5(P-1)+S
\end{equation}
where $P-1$ accounts for the baseline modification inherent in any transformation (e.g., selecting the target expression).

\textit{Strategy balancing.} Based on the calculated scores, we categorize strategies into three levels and apply differentiated filtering, detailed in Appendix~\ref{sec: appendix-strategy}. For \textit{easy} strategies (score$<1$), we remove all instances from the multiple-answer dataset, whereas for the single-answer version, we retain only $20\%$ of simplification-oriented strategies (e.g., log simplification) 
and a mere $4\%$ of their inverse expansion counterpart (e.g., expand log simplification). These different ratios are based on our observation that LLMs can typically generalize to expansion tasks after mastering the corresponding simplification logic. 
For \textit{medium} strategies ($1\leq$score$<2.5$), we solely cap their occurrences at 2,000 in the multi-answer version. 
Finally, all \textit{difficult} strategies (score $\geq$ 2.5) are fully retained to maximize the LLM's exposure to complex optimization logic.

\noindent \textbf{Dataset Summary.} Following the verification and filtering stages, the 35,878 entries initially generated during the strategy-driven phase were refined to a final collection of 24,953 high-quality samples. The curated dataset includes 7,537 single-answer instances and 17,416 multi-answer instances. In the final distribution (treating each strategy in multi-answer samples independently), difficult strategies account for the majority at 87.32$\%$, while medium and easy strategies constitute 12.17$\%$and 0.51$\%$, respectively. This composition prioritizes high-complexity reasoning while retaining a minimal baseline to consolidate and activate the model’s existing knowledge of fundamental optimizations.

\section{Evaluation}

 






In this section, we present an experimental study of LEIR and the Step-TP dataset. Some technical details are deferred to the appendix. The goal of the experimental study is to empirically validate and answer the following main questions for our design.

1. Can LEIR achieve better token efficiency than CUDA and TIR?

2. Can Step-TP enable executable, grounded transformations across a diverse set of strategies?

3. Can Step-TP support long-horizon optimization and exhibit generality across different GPUs?

\noindent \textbf{Overview.} The results answer these questions affirmatively, validating the effectiveness of LEIR and Step-TP.

\subsection{Experimental Setup}
\textbf{Testbed.} All experiments are conducted on a machine with 1536GB of host memory and eight NVIDIA H20-3e (140GB memory each). Unless otherwise specified, this platform serves as the default environment. To further evaluate hardware adaptability, we also perform some evaluations on a machine with eight NVIDIA A100 GPUs (80GB memory each) and 800GB of host memory.

\noindent \textbf{Models.} The experiments are conducted on a range of Qwen3 models with different parameter scales, including Qwen3-1.7B, Qwen3-8B, Qwen3-14B, and Qwen3-32B. 
The training setups are detailed in Appendix~\ref{sec:appendix-eval}

\subsection{Token Efficiency}

\begin{figure}[!t]
 \centering
   \includegraphics[width=0.38\textwidth]{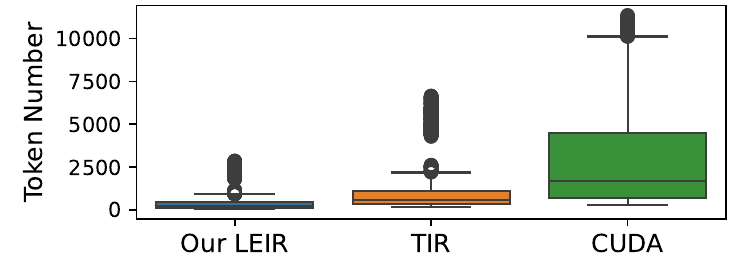}
 \caption{Token consumption of 6335 tensor programs across LEIR, TIR, and CUDA.
 }
 \label{fig:token-efficiency}
 \vspace{-6mm}
\end{figure}

To evaluate the context-efficiency of different IRs, we compare the token consumption of 6335 tensor programs across our LEIR, TVM TIR, and CUDA using the Qwen3 model's AutoTokenizer. As illustrated in Figure~\ref{fig:token-efficiency}, LEIR consistently exhibits the highest structural density. Specifically, the mean token count for LEIR is $499.3$, which is significantly lower than that of TVM TIR ($1244.2$) and CUDA ($2897.3$). The disparity is even more pronounced at the upper tail of the distribution: while the longest CUDA kernel consumes over $11,000$ tokens and risks exhausting the effective context window of many models, the maximum length for the equivalent LEIR representation remains under $2,900$ tokens. 

These results demonstrate that LEIR reduces the average token footprint by approximately $60\%$ compared to TIR and $83\%$ compared to CUDA. This efficiency ensures that complex optimization processes can be encoded within prompt-length limits, allowing the model to focus its reasoning capacity on strategy selection rather than parsing redundant syntax.

\subsection{Single-step Transformations}


\begin{table}[t]
\centering
\small
\begin{tabular}{lcccc} 
\hline
Model & Diff Pass& Build Pass & Exec Pass & Equal Pass\\\hline
Qwen3-1.7B& 99\%&87\%&87\%&73\%\\\hline
Qwen3-8B& 100\%&95\%&94\%&88\%\\\hline
Qwen3-14B& 100\%&96\%&96\%&90\%\\\hline
Qwen3-32B& 100\%&96\%&95\%&92\%\\\hline
\end{tabular}
\caption{Single-attempt pass rates on single-step tensor program transformations for models trained on StepTP.}
\label{tab:single-step-pass}
\vspace{-10mm}
\end{table}

\begin{figure}[!t]
 \centering
\subfigure[Category]{
   \includegraphics[width=0.225\textwidth]{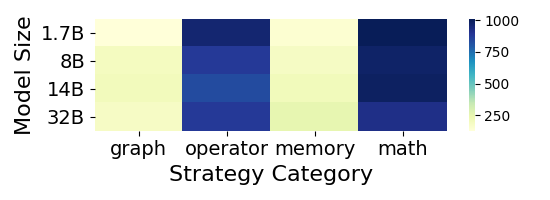}}
   \subfigure[Dfficulty]{
   \includegraphics[width=0.225\textwidth]{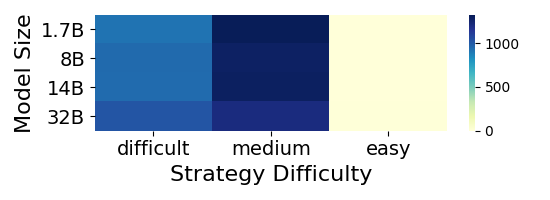}}
   \vspace{-4mm}
 \caption{Number of strategies applied by LLMs of different sizes across 2248 test cases, grouped by (a) strategy category and (b) difficulty level.
 }
 \label{fig:single-step-strategy}
 \vspace{-6mm}
\end{figure}

\begin{table*}[t]
\centering
\small
\begin{tabular}{l|cc|ccc|c} 
\hline
 Method&Avg. \# Samples& Max. \# Samples & Avg. Speedup & Median Speedup & Max. Speedup& Search Efficiency\\\hline
  Greedy Search &35.91&41& 20.79&	1.78&	193.55&0.58\\
 Beam Search&103.17&114&42.90&4.57&561.82&0.60\\
 BFS Search & 28.27&	31& 18.27&	2.11&	173.13& 0.65\\
 DFS Search & 28.11&	31& 20.97&	1.68&	242.62&0.75\\
 MCTS & 17.41&21& 24.96&	4.54&	171.9&1.43\\
 Chain Search & 16.63	&21& 23.62&	2.05&286.92&1.42\\
 Chain Search wo Parent&15.95&	21& 20.25&	4.96&	96.14&1.63\\\hline
 Chain Search on A100& 16.85	&21& 22.01&3.62&	265.35&1.31\\\hline
\end{tabular}
\caption{Performance comparison and search efficiency of various search algorithms guided by Qwen3-32B trained on Step-TP.}
\label{tab:multi-step-result}
\vspace{-6mm}
\end{table*}

\begin{table}[t]
\centering
\small
\begin{tabular}{lcc} 
\hline
 Method&Avg. \# Strategy & Max. \# Strategy \\\hline
  Greedy Search &3.6&8\\
 Beam Search&5.23&10\\
 BFS Search &2.45&4\\
 DFS Search & 2.49&4\\
 MCTS & 3.65&9\\
 Chain Search & 3.57&7\\
 Chain Search wo Parent&3.43&7\\\hline
 Chain Search on A100& 3.33&9\\\hline
\end{tabular}
\caption{Number of strategies of multi-step optimization trajectories guided by Qwen3-32B trained on Step-TP.}
\label{tab:multi-step-strategy}
\vspace{-10mm}
\end{table}

We evaluate models trained on our dataset via single-step transformations, highlighting two key dataset properties: (i) enabling faithfully grounded, executable program transformations, and (ii) preliminarily supporting a diverse set of tensor-program-level strategies.

\noindent \textbf{Setup.} We evaluate all models trained on our dataset across 2248 test cases, constructed from 180 distinct tensor programs by varying input/output data types and shapes. All test cases are strictly held out from training. 
Each model is prompted using our single-answer format, requiring it to autonomously select a valid transformation.

\noindent \textbf{Metrics.} We employ four progressively stricter metrics to evaluate the transformations, where each success is counted only after passing three independent verification trials to ensure stability: (i) Different pass, to ensure the generated LEIR is syntactically modified; (ii) Build pass, to confirm the transformed LEIR is compilable into CUDA; (iii) Execute pass, to verify the successful execution; and (iv) Equal pass, to validate equivalence with the original program.  

\noindent \textbf{Result Analysis.} Table~\ref{tab:single-step-pass} demonstrates that models fine-tuned on our Step-TP dataset exhibit exceptional fidelity to executable and equivalent program transformations. Specifically, all models achieve near-perfect different pass rates, 
with Qwen3-32B reaching a 92\% equal pass rate and even the 1.7B model achieving 73\%. 
The consistently high success rates in build and execute passes suggest that our reasoning traces effectively guide the models to maintain functional correctness during complex IR modifications. 
To further examine the effect of repeated sampling, we also evaluate Step-TP with more independent attempts in Appendix~\ref{sec:appendix-eval}. Qwen3-8B trained on our dataset improves from 88.08\% equal Pass with a single attempt to 95.55\% with 10 attempts and 96.22\% with 16 attempts.
This underscores the superior quality and robust generalization of the Step-TP dataset.

We further analyze the diversity of strategies autonomously selected by the models. 
As shown in Figure~\ref{fig:single-step-strategy}(a), the category-wise distribution (graph, operator, memory, math) closely mirrors the underlying dataset ratio (8:9:5:21), confirming that models successfully cover all strategy categories proportionally.
As shown in Figure~\ref{fig:single-step-strategy}(b), easy strategies occur least frequently, and medium and difficult strategies appear at comparable rates
indicating that the models do not avoid difficult strategies in favor of simpler alternatives. 
Moreover, we observe that the 32B model utilizes 31 unique strategies at all levels, while other scales cover 29, representing over 70\% of the 43 total available strategies. This diverse coverage, aligned with our data distribution, confirms that the models have successfully mastered a broad spectrum of representative optimization patterns.

\subsection{Multi-step Optimizations}

We evaluate models trained on our dataset via multi-step transformations, highlighting three key dataset properties: (i) supporting long-horizon complex optimization by combining multiple strategies step by step, (ii) enabling efficient search through high-quality step-level supervision, and (iii) exhibiting generality across different GPU environments. We provide a detailed case study of the highest-performing test case in Appendix~\ref{sec:appendix-eval}.

\noindent \textbf{Setup.} We evaluate Qwen3-32B, trained on Step-TP, across 100 distinct tensor programs strictly held out from training. The model is tasked with generating runtime-performance-optimized, equivalent LEIRs.
To implement multi-step optimization, we deploy seven distinct search algorithms: Greedy Search, Breadth-First Search (BFS), Depth-First Search (DFS), Beam Search, Monte Carlo Tree Search (MCTS), and Chain-based Search (with/without parent nodes). 
The detailed setups are provided in Appendix~\ref{sec:appendix-eval}.


\noindent \textbf{Metrics.} We evaluate results using four metrics: (i) \# Samples: the total number of candidate LEIRs verified during the entire optimization process; (ii) Speedup: the runtime improvement of the transformed LEIR relative to the original LEIR 
, computed as $\text{Runtime}_\text{original}/\text{Runtime}_\text{transformed}$; (iii) \# Strategies: the count of distinct strategy types applied along the final optimization trajectory, where each type is counted once regardless of different applications 
; and (iv) Search efficiency: the average speedup achieved per verified sample, calculated as $\text{Average Speedup}/\text{\# Samples}$.

\noindent \textbf{Result Analysis.} We evaluate the ability of our dataset Step-TP across all search algorithms. Table~\ref{tab:multi-step-result} details the number of samples, the speedup, and the search efficiency, while Table~\ref{tab:multi-step-strategy} tracks the structural complexity of the optimization trajectories via the number of strategies. The results illustrate key properties:

\textbf{\textit{(i) Supporting Long-horizon Complex Optimization.}} 
As shown in Table \ref{tab:multi-step-result}, across all search algorithms, the model trained on Step-TP consistently achieves substantial performance improvements across all search paradigms. For instance, Beam Search achieves an average speedup of 42.90$\times$ and a peak speedup of 561.82$\times$. Notably, even under the most restrictive search budgets (e.g., Chain-based Search without parent nodes), the model maintains a high median speedup of 4.96$\times$. These results indicate performance gains are not driven by isolated outliers but demonstrate that our dataset captures high-quality optimization patterns, enabling algorithms to successfully compose sequences of transformations.

Table~\ref{tab:multi-step-strategy} further reveals that optimal schedules require an average of 2.45 to 5.23 distinct strategy types, with a maximum depth of 10 (e.g., in Beam Search). This demonstrates the support of our dataset for diverse tensor-program-level strategies, which serve as the essential building blocks for composing these optimal trajectories.

Together, these observations confirm that our dataset provides the necessary structural knowledge to navigate non-trivial, long-horizon optimization landscapes.

\textbf{\textit{(ii) Enabling Efficient Search.}} 
As shown in Table~\ref{tab:multi-step-result}, the search efficiency ranging from 0.58 to 1.63 indicates that the majority of verified transformations contribute meaningfully to the final speedup.
In particular, even search algorithms without backtracking or branching techniques achieve significant performance. For instance, Greedy Search and Chain-based Search variants achieve average speedups of $20.25\times$ to $25.97\times$ while requiring as few as 16 to 36 samples. The success of these short-insight paradigms suggests that Step-TP provides high-quality step-level supervision that enables the model to effectively identify high-potential transformation paths, reducing the cost of exhaustive trial and error.

\textit{(iii) Exhibiting generality across different GPUs.} As shown in Tables~\ref{tab:multi-step-result} and \ref{tab:multi-step-strategy}, Chain-based Search on A100 GPU maintains a strong speedup performance and complex trajectories up to 9 strategies, extending beyond the default H20-3e GPUs. This confirms that the optimization knowledge captured by Step-TP 
remains effective across different GPU generations.


\subsection{Ablation Study}
\label{sec:ablation}

We conduct ablation studies to examine how three key Step-TP dataset designs affect the performance of trained models: the LEIR representation, structured CoT supervision, and strategy filtering.

\noindent \textbf{Setup.}
Four controlled dataset variants are constructed from the same source programs and transformation pipeline. Specifically, we train Qwen3-8B on: (i) TensorIR-based data without CoT but with strategy filtering (\textit{F TIR wo CoT}); (ii) LEIR-based data without CoT but with strategy filtering (\textit{F LEIR wo CoT}); (iii) LEIR-based data with CoT but without strategy filtering (\textit{UF LEIR with CoT}); and (iv) the full Step-TP dataset with all three components enabled. 

\begin{table}[t]
\centering
\small
\setlength{\tabcolsep}{5pt}
\begin{tabular}{lcc}
\hline
Dataset Variant & Equal Pass & Difficult:Medium:Easy strategy Ratio \\
\hline
F TIR wo CoT  & 77\% & 42:51:7 \\
F LEIR wo CoT & 83\% & 44:50:6 \\
UF LEIR with CoT & 88\% & 25:28:47 \\
Step-TP                 & 88\% & 42:51:7 \\
\hline
\end{tabular}
\caption{Single-step transformation ablation results for Qwen3-8B trained on four dataset variants on 2248 test cases.}
\label{tab:ablation-single}
\vspace{-4mm}
\end{table}

\begin{table}[t]
\centering
\small
\setlength{\tabcolsep}{4pt}
\begin{tabular}{lccc}
\hline
Dataset Variant & Avg. \#Samples & Avg. Speedup & Search Efficiency \\
\hline
F TIR wo CoT  & 8.10  & 0.72  & 0.08 \\
F LEIR w/o CoT & 7.31  & 0.75  & 0.10 \\
UF LEIR with CoT & 13.66 & 8.14  & 0.59 \\
Step-TP                 & 13.88 & 12.10 & 0.87 \\
\hline
\end{tabular}
\caption{Multi-step optimization ablation results for Qwen3-8B trained on four dataset variants on 100 distinct test cases using chain-based search with parent nodes.}
\label{tab:ablation-multi}
\vspace{-4mm}
\end{table}

\noindent \textbf{Result Analysis.}
We compare the four dataset variants from two perspectives: single-step transformations, which measure transformation fidelity and strategy difficulty, and multi-step optimization, which measures optimization effectiveness and search efficiency. Table~\ref{tab:ablation-single} reports single-step transformation performance in terms of Equal Pass and the difficulty distribution of the applied strategies, while Table~\ref{tab:ablation-multi} reports the average number of verified samples, average speedup, and search efficiency during multi-step optimization. These results lead to the following observations:


\textit{(i) LEIR reduces representation-induced transformation errors.}
As shown in Table~\ref{tab:ablation-single}, replacing TensorIR with LEIR improves equal pass from 77\% to 83\% when CoT is removed and strategy filtering is kept. This suggests that LEIR helps the model generate semantically equivalent transformations by exposing the relevant loop and equation structures more directly, instead of requiring the model to reason through compiler-oriented TensorIR boilerplate.

\textit{(ii) Structured CoT is critical for long-horizon optimization.}
In single-step transformation, Table~\ref{tab:ablation-single} shows that adding CoT improves equal pass from 83\% to 88\%, indicating better semantic preservation. In multi-step optimization, Table~\ref{tab:ablation-multi} shows a much larger improvement, with average speedup increasing from 0.75 to 12.10 and search efficiency from 0.10 to 0.87. This suggests that structured CoT helps the model learn composable transformation logic, enabling effective optimization trajectories beyond locally valid IR edits.

\textit{(iii) Strategy filtering reduces the bias toward simplistic transformations.}
As shown in Table~\ref{tab:ablation-single}, adding strategy filtering keeps equal pass unchanged at 88\%, but changes the generated strategy distribution from 25:28:47 to 42:51:7 for difficult, medium, and easy strategies, respectively. This indicates that filtering suppresses the model's preference for easy transformations without sacrificing single-step correctness. This shift becomes more meaningful in multi-step optimization, where Table~\ref{tab:ablation-multi} shows higher average speedup from 8.14 to 12.10 and higher search efficiency from 0.59 to 0.87. 


\section{Conclusion}
We introduce Step-TP, a step-level post-training dataset for LLM-based tensor program optimization that provides verifiable, compositional supervision for single-step transformation decisions. By combining a token-efficient intermediate representation (LEIR) with atomic strategy decomposition and deterministic equivalence checking, Step-TP enables models to reason about precise optimization steps, rather than relying on outcome-only shortcuts.
\section*{ACKNOWLEDGEMENT}
This work was supported in part by a collaborative research grant from Ant Group and grants from Hong Kong RGC under the contracts 17204423, 17205824, 17204625, C7004-22G (CRF), CRS\_PolyU501/23, and T43-513/23-N (TRS).

\bibliographystyle{ACM-Reference-Format}
\bibliography{sample-base}
\newpage
\clearpage
\appendix
\section{Intermediate representation}\label{sec:appendix-IR}

\textbf{\textit{Loop structure.}} The loop structure is represented by a main symbol indicating the loop type, with superscripts and subscripts specifying the loop index and iteration range.

(1). Main symbol:  Five loop notations represent specific loop types: serial loops ($L$), parallel loops ($P$), vectorized loops ($V$), unrolled loops ($U$), and thread/block-binding loops ($B$). 

(2). Superscript and subscript: The numbers in the superscript and subscript denote the loop range, such as $L^{32}_{a=0}$ for a range of $[0, 32)$. The lowercase letters in the subscript indicate the loop index. Specifically, indices for thread/block-binding loops are categorized into six mappings to CUDA intrinsics: $\{bx, by, bz\}$ for \texttt{blockIdx} and $\{tx, ty, tz\}$ for \texttt{threadIdx}. For example, $B^{6}_{tx=0}$ denotes a loop bound to \texttt{threadIdx.x}.

\textbf{\textit{Equation structure.}} The equation structure consists of three components: the element-wise computation, the variables to denote tensors, and delimiters to define the computational scope and logical sequence within the loop domain. Furthermore, we employ default initialization to maintain an uninterrupted algebraic flow.

\textit{Element-wise computation.} We formalize the algebraic computations using a set of functional operators derived from TVM TIR, including standard arithmetic (e.g., $+$, $-$), transcendental functions (e.g., $\exp, \log$), and conditional intrinsics (e.g., $\text{if\_then\_else}$), all applied in a purely element-wise manner.

\textit{Variable.} Each tensor variable is defined by a main symbol for its identity, with subscripts specifying indices and superscripts indicating data and memory type.

(1). Main symbol: Tensor variables (input, output, and intermediate) are represented by an uppercase letter followed by optional lowercase letters (e.g., $A, Ac, Cem$). To avoid semantic ambiguity with loop types and existing compiler namespaces (such as the TVM T package), the symbols $\{L, P, V, U, B, T\}$ are reserved and excluded from variable naming.

(2). Subscript: The subscript specifies the access indices using lowercase letters. To maintain semantic clarity and index integrity, certain characters are reserved: $\{t, b, x, y, z\}$ are dedicated to loop binding semantics, while $e$ is excluded to avoid syntactic confusion with the exponential operator ($\exp$). 

(3). Superscript: The superscript encodes two mandatory attributes: the data type (e.g., $f32$, $i64$, $f16$ for FLOAT32, INT64, FLOAT16) and the memory type ($g$ for global, $l$ for local, $s$ for shared memory).

To ensure a fully specified execution state, both subscripts and superscripts are mandatory for all tensor variables (e.g., $Ac^{f32,g}_{a,c,e}$). The only exception is for scalars, where subscripts are omitted as they lack associated iteration axes.

\textit{Delimiter.} Delimiters explicitly define the scope of computations and separate independent expressions within the IR.  
 
(1). Brackets: Square brackets $[$ and $]$ enclose one or more equations to define their computational scope, indicating that the enclosed logic is executed under the specified loop conditions. For example, For example, $B^{8}_{tx=0}L^{16}_{a=0}[C^{f32,s}_{tx*16+a}=As^{f32,s}_{tx*16+a}+1;];$ denotes the element-wise computation under a nested structure consisting of a thread-binding loop $tx$ and a serial loop $a$.
 
(2). Semicolon: The semicolon $;$ acts as a separator for both independent equations and top-level expressions, where an expression comprises the loop nest and its enclosed equations. As illustrated in the example, $L^{4}_{a=0}[C^{f32,g}_{a}= A^{f32,g}_{a}*3; D^{f32,g}_{a}= A^{f32,g}_{a}*2+1;]; L^{4}_{a=0}[H^{f32,g}_{a}= A^{f32,g}_{a}*2;];$, the semicolon enforces a sequential execution order: (i) within a loop body, it separates distinct equations (e.g., to obtain $C$ and $D$), ensuring their sequential evaluation within every iteration. (ii) At the expression level, it separates independent loop nests, indicating that the second loop (e.g., to calculate $H$) begins execution only after the entire preceding loop nest has completed.

\textit{Default Initialization.} To enhance brevity and semantic density, our IR employs implicit initialization for common reduction patterns, including summation, product, and extremum operations. The initial identity element is automatically inferred from the operator: summations default to $0$, products to $1$, and max or min reductions to $-\infty$ or $+\infty$, respectively. 
\section{Strategy}\label{sec: appendix-strategy}

\subsection{Tensor-Program-Level Strategies}

\begin{table*}[t]
\centering
\begin{tabular}{p{0.3\textwidth} p{0.6\textwidth}} 
\hline
Strategy & Description and Examples\\\hline
operator fusion &  it places consecutive equations from multiple similar loop nests into a single loop nest. For example, the consecutive expressions $B^{478}_{tx=0}L^{478}_{a=0}[E^{f16,g}_{tx,a}=\text{if\_then\_else}(tx>=a,C^{f16,g}_{tx,a},0);];$ and $B^{478}_{tx=0}L^{478}_{a=0}[D^{f16,g}_{tx,a}=\text{if\_then\_else}((tx>=a,A^{f16,g}_{tx,a},0);];$ has the same loop nest $B^{478}_{tx=0}L^{478}_{a=0}$, and are fused into the expression $B^{478}_{tx=0}L^{478}_{a=0}[D^{f16,g}_{tx,a}=\text{if\_then\_else}((tx>=a,A^{f16,g}_{tx,a},0);E^{f16,g}_{tx,a}=\text{if\_then\_else}((tx>=a,C^{f16,g}_{tx,a},0);];$.
\\\hline
operator fission & it splits multiple equations in one loop nest into multiple separate loop nests. For example, $B^{975}_{tx=0}L^{10081}_{a=0}[C^{f64,s}_{tx,a}=1/(1+exp(-A^{f64,g}_{tx,a}));D^{f64,s}_{tx,a}=A^{f64,g}_{tx,a}*C^{f64,s}_{tx,a};];$ with multiple equations are split into $B^{975}_{tx=0}L^{10081}_{a=0}[D^{f64,g}_{tx,a}=A^{f64,g}_{tx,a}*C^{f64,g}_{tx,a};];$ and $B^{975}_{tx=0}L^{10081}_{a=0}[C^{f64,g}_{tx,a}=1/(1+exp(-A^{f64,g}_{tx,a}));];$ as two expressions.\\\hline
compute inline & it merges related equations from multiple loop nests into one equation within a single loop nest. For example, the expressions $B^{175}_{tx=0}L^{28272}_{a=0}[C^{f16,g}_{tx,a}=abs(A^{f16,g}_{tx,a});];$ and $B^{175}_{tx=0}L^{28272}_{a=0}[F^{f16,g}_{tx,0}=F^{f16,g}_{tx,0}+C^{f16,g}_{tx,a};];$ has the same loop nest $B^{175}_{tx=0}L^{28272}_{a=0}$ and related equations, and are merged into the expression $B^{175}_{tx=0}L^{28272}_{a=0}[F^{f16,g}_{tx,0}=F^{f16,g}_{tx,0}+abs(A^{f16,g}_{tx,a});];$.
\\\hline
expression splitting & it separates a merged equation in one loop nest into multiple equations under multiple loop nests. For example, $B^{929}_{tx=0}L^{10637}_{a=0}[E^{f32,g}_{tx,a}=exp(D^{f32,g}_{tx,a}-F^{f32,g}_{tx})/G^{f32,g}_{tx};];$ with the merged equation are split into $B^{929}_{tx=0}L^{10637}_{a=0}[J^{f32,g}_{tx,a}=exp(D^{f32,g}_{tx,a}-F^{f32,g}_{tx});]; and B^{929}_{tx=0}L^{10637}_{a=0}[E^{f32,g}_{tx,a}=J^{f32,g}_{tx,a}/G^{f32,g}_{tx};];$ as two expressions
\\\hline
tensor concat to fuse operators & it concatenates multiple input variables into one variable, merges the equations, and then splits the output variable to obtain multiple outputs. For example, for two similar expressions $B^{99}_{tx=0}L^{1167}_{a=0}L^{450}_{c=0}[D^{f32,g}_{tx,a,c}=exp(A^{f32,g}_{tx,a,c});];$ and $B^{99}_{tx=0}L^{450}_{a=0}L^{365}_{c=0}[E^{f32,g}_{tx,a,c}=exp(C^{f32,g}_{tx,a,c});];$, the input variables are contenated by $B ^{99}_{tx=0}L^{1167}_{a=0}L^{450}_{c=0}[G^{f32,g}_{tx,a,c}=A^{f32,g}_{tx,a,c};];$ and $B^{99}_{tx=0}L^{450}_{a=0}L^{365}_{c=0}[G^{f32,g}_{tx,a+1167,c+450}=C^{f32,g}_{tx,a,c};];$, the similar operations are executed by $B^{99}_{tx=0}L^{1617}_{a=0}L^{815}_{c=0}[H^{f32,g}_{tx,a,c}=exp(G^{f32,g}_{tx,a,c});];$, and the output variable is split into two outputs by $B^{99}_{tx=0}L^{1167}_{a=0}L^{450}_{c=0}[D^{f32,g}_{tx,a,c}=H^{f32,g}_{tx,a,c};];$ and $B^{99}_{tx=0}L^{450}_{a=0}L^{365}_{c=0}[E^{f32,g}_{tx,a,c}=H^{f32,g}_{tx,a+1167,c+450};];$
\\\hline
tensor split to decouple operators & it splits an input variable into multiple variables, runs multiple equations, and then concatenates multiple output variables into one variable. For example, for the expression $B^{687}_{tx=0}L^{217}_{a=0}[C^{f32,g}_{tx,a}=max(0,min(1,(A^{f32,g}_{tx,a}+3)/6));];$, the input variable is split into two inputs by $B^{57}_{tx=0}L^{217}_{a=0}[C^{f32,g}_{tx,a}=F^{f32,g}_{tx,a};];$ and $B^{630}_{tx=0}L^{217}_{a=0}[E^{f32,g}_{tx,a}=A^{f32,g}_{tx+57,a};];, two similar operations are executed by B^{630}_{tx=0}L^{217}_{a=0}[C^{f32,g}_{tx+57,a}=G^{f32,g}_{tx,a};];$ and $B^{57}_{tx=0}L^{217}_{a=0}[D^{f32,g}_{tx,a}=A^{f32,g}_{tx,a};];$, and two output variables are concatenated into one output by $B^{57}_{tx=0}L^{217}_{a=0}[F^{f32,g}_{tx,a}=max(0,min(1,(D^{f32,g}_{tx,a}+3)/6));];$ and $B^{630}_{tx=0}L^{217}_{a=0}[G^{f32,g}_{tx,a}=max(0,min(1,(E^{f32,g}_{tx,a}+3)/6));];$.
\\\hline
common subexpression elimination & it computes duplicated expressions once and reuses the result to avoid redundant calculations. For example, a common equation part $exp(A^{f64,g}_{tx,a})$ exists, so this part can be computed in the expression $B^{917}_{tx=0}L^{30201}_{a=0}[F^{f64,g}_{tx,a}=exp(A^{f64,g}_{tx,a});];$ once.
\\\hline
expression reorder & it rearranges the expressions. For example, two expressions $B^{448}_{tx=0}L^{2}_{a=0}L^{549}_{c=0}L^{549}_{d=0}L^{154}_{f=0}[Q^{f64,g}_{tx,a,c,d}=Q^{f64,g}_{tx,a,c,d}+K^{f64,g}_{tx,a,c,f}*O^{f64,g}_{tx,a,f,d};];$ and  $B^{1}_{tx=0}[R^{f64,g}=154;];$ can be reordered.
\\
\hline
\end{tabular}
\caption{Graph-level atomic optimization strategies.}
\label{tab:graph-level-strategy}
\end{table*}

\begin{table*}[t]
\centering
\begin{tabular}{p{0.3\textwidth} p{0.6\textwidth}} 
\hline
Strategy & Description and Example\\\hline
loop reorder& it rearranges the nesting orders of loops within a loop nest. For example, the loops $B^{55}_{tx=0}L^{277}_{a=0}L^{4}_{c=0}L^{124}_{d=0}$ in the expression $B^{55}_{tx=0}L^{277}_{a=0}L^{4}_{c=0}L^{124}_{d=0}[Ad^{f16,g}_{tx,a,c,d}=Ac^{f16,g}_{tx,a,c,d};];$ can be reordered as the loops $B^{55}_{tx=0}L^{124}_{d=0}L^{4}_{c=0}L^{277}_{a=0}$ in the expression $B^{55}_{tx=0}L^{124}_{d=0}L^{4}_{c=0}L^{277}_{a=0}[Ad^{f16,g}_{tx,a,c,d}=Ac^{f16,g}_{tx,a,c,d};];$.
\\\hline
loop tiling& The strategy loop tiling is used on the given IR to break two nested loops into four smaller loops (i.e., forming the tiles) within one loop nest. For example, the loops $B^{255}_{tx=0}L^{255}_{a=0}L^{255}_{c=0}$ in the expression $B^{255}_{tx=0}L^{255}_{a=0}L^{255}_{c=0}[D^{f16,g}_{tx,a}=D^{f16,g}_{tx,a}+A^{f16,g}_{tx,c}*C^{f16,g}_{c,a};];$ can be tiled as the loops $L^{15}_{g=0}L^{3}_{h=0}B^{17}_{tx=0}L^{85}_{a=0}L^{255}_{c=0}$ in the expression $L^{15}_{g=0}L^{3}_{h=0}B^{17}_{tx=0}L^{85}_{a=0}L^{255}_{c=0}[D^{f16,g}_{g*17+tx,h*85+a}=D^{f16,g}_{g*17+tx,h*85+a}+A^{f16,g}_{g*17+tx,c}*C^{f16,g}_{c,h*85+a};];$.
\\\hline
loop split& it divides any loop within a loop nest into two nested loops. For example, the loops $B^{16}_{tx=0}L^{16384}_{a=0}$ in the expression $B^{16}_{tx=0}L^{16384}_{a=0}[E^{f16,g}_{tx,a}=A^{f16,g}_{tx,a}/D^{f16,g}_{tx,a};];$ can be split as the loops $L^{4}_{f=0}B^{4}_{tx=0}L^{16384}_{a=0}$ in the expression $L^{4}_{f=0}B^{4}_{tx=0}L^{16384}_{a=0}[E^{f16,g}_{f*4+tx,a}=A^{f16,g}_{f*4+tx,a}/D^{f16,g}_{f*4+tx,a};];$
\\\hline
loop fusion& it combines multiple loops within a loop nest into one loop. For example, the loops $L^{2}_{f=0}B^{114}_{tx=0}L^{491}_{a=0}$ in the expression $L^{2}_{f=0}B^{114}_{tx=0}L^{491}_{a=0}[E^{f32,g}_{a,f*114+tx}=C^{f32,g}_{f*114+tx,a};];$ can be fused as the loops $B^{228}_{tx=0}L^{491}_{a=0}$ in the expression $B^{228}_{tx=0}L^{491}_{a=0}[E^{f32,g}_{a,tx}=C^{f32,g}_{tx,a};];$.
\\\hline
loop unrolling& it expands a loop body by computing its equation multiple times. For example, one loop axis in the loops $B^{16}_{tx=0}L^{32}_{a=0}L^{64}_{c=0}L^{64}_{d=0}L^{2}_{f=0}L^{2}_{g=0}$ of the expression $B^{16}_{tx=0}L^{32}_{a=0}L^{64}_{c=0}L^{64}_{d=0}L^{2}_{f=0}L^{2}_{g=0}[C^{f64,g}_{tx,a,c,d}=max(C^{f64,g}_{tx,a,c,d},D^{f64,g}_{tx,a,c*2+f*3,d*2+g*3});];$ can be unrolled, then the loops become $B^{16}_{tx=0}U^{32}_{a=0}L^{64}_{c=0}L^{64}_{d=0}L^{2}_{f=0}L^{2}_{g=0}$ of the expression $B^{16}_{tx=0}U^{32}_{a=0}L^{64}_{c=0}L^{64}_{d=0}L^{2}_{f=0}L^{2}_{g=0}[C^{f64,g}_{tx,a,c,d}=max(C^{f64,g}_{tx,a,c,d},D^{f64,g}_{tx,a,c*2+f*3,d*2+g*3});];$.
\\\hline
loop parallelization& it distributes iterations of a loop across multiple processing units. For example, one loop axis in the loops $B^{851}_{tx=0}L^{36}_{a=0}L^{979}_{c=0}L^{36}_{d=0}$ of the expression 
$B^{851}_{tx=0}L^{36}_{a=0}L^{979}_{c=0}L^{36}_{d=0}[E^{f32,g}_{tx,a,c,d}=\text{if\_then\_else}(1<=c<978\&2<=d<34,A^{f32,g}_{tx,a,c-1,d-2},0);];$ can be parallel, then the loops become $B^{851}_{tx=0}L^{36}_{a=0}P^{979}_{c=0}L^{36}_{d=0}$ of the expression $B^{851}_{tx=0}L^{36}_{a=0}P^{979}_{c=0}L^{36}_{d=0}[E^{f32,g}_{tx,a,c,d}=\text{if\_then\_else}(1<=c<978\&2<=d<34,A^{f32,g}_{tx,a,c-1,d-2},0);];$
\\\hline
loop vectorization& it transforms loop operations to use SIMD instructions and thereby processes multiple data elements simultaneously. For example, one loop axis in the loops $B^{178}_{tx=0}L^{105}_{a=0}L^{1}_{c=0}$ of the expression $B^{178}_{tx=0}L^{105}_{a=0}L^{1}_{c=0}[F^{f64,g}_{tx,a,c,0}=E^{f64,g}_{tx,a,c};];$ can be vectorized, then the loops become $B^{178}_{tx=0}L^{105}_{a=0}V^{1}_{c=0}$ of the expression $B^{178}_{tx=0}L^{105}_{a=0}V^{1}_{c=0}[F^{f64,g}_{tx,a,c,0}=E^{f64,g}_{tx,a,c};];$
\\\hline
loop binding& it maps loop iterations to specific GPU threads or blocks along [x, y, z] axes. For example, one loop axis in the loops $B^{595}_{tx=0}L^{71}_{a=0}L^{595}_{c=0}$ of the expression $B^{595}_{tx=0}L^{71}_{a=0}L^{595}_{c=0}[E^{f16,g}_{tx,a}=E^{f16,g}_{tx,a}+D^{f16,g}_{tx,c}*C^{f16,g}_{c,a};];$ can be binded, then the loops become $B^{595}_{tx=0}B^{71}_{bza=0}L^{595}_{c=0}$ of the expression $B^{595}_{tx=0}B^{71}_{bza=0}L^{595}_{c=0}[E^{f16,g}_{tx,bza}=E^{f16,g}_{tx,bza}+D^{f16,g}_{tx,c}*C^{f16,g}_{c,bza};];$.
\\\hline
reduction factorization& it restructures a reduction expression into multiple reduction expressions by splitting the reduction loop axis and then adds the outputs. For example, the reduction loop axis in the loops $B^{32}_{tx=0}L^{4}_{a=0}L^{2}_{c=0}L^{1186}_{d=0}L^{5875}_{f=0}$ of the expression $B^{32}_{tx=0}L^{4}_{a=0}L^{2}_{c=0}L^{1186}_{d=0}L^{5875}_{f=0}[F^{f32,g}_{tx,a}=F^{f32,g}_{tx,a}+A^{f32,g}_{tx,a*2+c,d,f};];$ can be split, so the loops become $B^{32}_{tx=0}L^{4}_{a=0}L^{1186}_{d=0}L^{5875}_{f=0}$ of the expression $B^{32}_{tx=0}L^{4}_{a=0}L^{1186}_{d=0}L^{5875}_{f=0}[F^{f32,g}_{tx,a}=F^{f32,g}_{tx,a}+(K^{f32,g}_{tx,a}+M^{f32,g}_{tx,a});];$ and $B^{32}_{tx=0}L^{4}_{a=0}L^{1}_{c=0}L^{1186}_{d=0}L^{5875}_{f=0}$ of the expression $B^{32}_{tx=0}L^{4}_{a=0}L^{1}_{c=0}L^{1186}_{d=0}L^{5875}_{f=0}[K^{f32,g}_{tx,a}=K^{f32,g}_{tx,a}+A^{f32,g}_{tx,a*2+c,d,f};];$. Then the outputs are added by $B^{32}_{tx=0}L^{4}_{a=0}L^{1}_{c=0}L^{1186}_{d=0}L^{5875}_{f=0}[M^{f32,g}_{tx,a}=M^{f32,g}_{tx,a}+A^{f32,g}_{tx,a*2+c+1,d,f};];$
\\
\hline
\end{tabular}
\caption{Operator-level atomic optimization strategies.}
\label{tab:operator-level-strategy}
\end{table*}

\begin{table*}[t]
\centering
\begin{tabular}{p{0.3\textwidth} p{0.6\textwidth}} 
\hline
Strategy & Description and Example\\\hline
cache read/write& it moves tensor variables between global (g), shared (s), and local (l) memory (i.e., shown in the superscript of the variable). For example, the variable $D^{f32,g}_{tx,a}$ in the expression $B^{413}_{tx=0}L^{892}_{a=0}[D^{f32,g}_{tx,a}=log(A^{f32,g}_{tx,a});];$ can write from local memory as $I^{f32,l}_{tx,a}$ by two expressions $B^{413}_{tx=0}L^{892}_{a=0}[D^{f32,g}_{tx,a}=I^{f32,l}_{tx,a};];$ and $B^{413}_{tx=0}L^{892}_{a=0}[I^{f32,l}_{tx,a}=log(A^{f32,g}_{tx,a});];$.
\\\hline
layout transformation& it changes the memory arrangement of tensor variables (i.e., shown in the subscript of the variable). For example, the memory layout of variable $A$ can be transformed as the variable $E$ by the expression $L^{11}_{g=0}B^{77}_{tx=0}L^{3182}_{c=0}[E^{f16,g}_{g,tx,c}=A^{f16,g}_{g*77+tx,c};];$
\\\hline
set storage scope& it directly sets intermediate tensor variables between global (g), shared (s), and local (l) memory (i.e., shown in the superscript of the variable). For example, the intermediate variable $E^{f32,g}_{tx}$ can be set in shared memory as $E^{f32,s}_{tx}$.
\\\hline
set storage layout& it directly changes the memory arrangement of intermediate tensor variables (i.e., shown in the subscript of the variable). For example, the memory layout of the intermediate variable $D$ can be directly set by updating its subscripts (e.g., from ${tx,a}$ to ${a*311+tx}$). Note: all occurrences of D in the current IR are updated consistently.
\\\hline
precompute indices& it precomputes indices of tensor variable (i.e., shown in the subscript of the variable) to store frequently used index expressions for reuse. For example, the indices c*3+g in the expression $B^{99}_{tx=0}L^{26}_{a=0}L^{36}_{c=0}L^{253}_{d=0}L^{109}_{f=0}L^{3}_{g=0}L^{1}_{h=0}[C^{f32,g}_{tx,a,c,d}=C^{f32,g}_{tx,a,c,d}+E^{f32,g}_{tx,f,c*3+g,d*3+h}*D^{f32,g}_{a,f,g,h};];$ can be precomputed as the variable $F^{i64,g}_{c,g}$ by the expression $B^{36}_{tx=0}L^{3}_{g=0}[F^{i64,g}_{tx,g}=tx*3+g;];$
\\
\hline
\end{tabular}
\caption{Memory-level atomic optimization strategies.}
\label{tab:memory-level-strategy}
\end{table*}

\begin{table*}[t]
\centering
\begin{tabular}{p{0.3\textwidth} p{0.6\textwidth}} 
\hline
Strategy & Description and Example\\\hline
factorization& it decomposes a mathematical equation into products or sums of simpler components. For example, the equation in the expression $B^{166}_{tx=0}L^{13601}_{a=0}[C^{f64,g}_{tx,a}=max(0,min(1,(A^{f64,g}_{tx,a}+3)/6));];$ can use the factorization strategy to transform as the expression $B^{166}_{tx=0}L^{13601}_{a=0}[C^{f64,g}_{tx,a}=max(0,min(1,A^{f64,g}_{tx,a}/6+1/2));];$.
\\\hline
expand factorization& it reconstructs a factored equation into its original full equation (i.e., the reverse process of factorization). For example, the equation in the expression $B^{7}_{tx=0}L^{3545}_{a=0}[G^{f16,g}_{tx}=G^{f16,g}_{tx}+(A^{f16,g}_{tx,a}-D^{f16,g}_{tx,a})**2;];$ can use the expand factorization strategy to transform as the expression $B^{7}_{tx=0}L^{3545}_{a=0}[G^{f16,g}_{tx}=G^{f16,g}_{tx}+A^{f16,g}_{tx,a}**2-2*A^{f16,g}_{tx,a}*D^{f16,g}_{tx,a}+D^{f16,g}_{tx,a}**2;];$
\\\hline
cancellation& it removes variables that offset each other to simplify the equation. For example, the equation in the expression $B^{128}_{tx=0}L^{16}_{a=0}L^{30}_{c=0}L^{30}_{d=0}[D^{f64,g}_{tx,a,c,d}=H^{f64,g}_{a}*((C^{f64,g}_{tx,a,c,d}-K^{f64,g}_{tx,a})/sqrt(N^{f64,g}_{tx,a}+1e-05))+I^{f64,g}_{a};];$ can use the cancellation strategy to transform as the expression $B^{128}_{tx=0}L^{16}_{a=0}L^{30}_{c=0}L^{30}_{d=0}[D^{f64,g}_{tx,a,c,d}=(C^{f64,g}_{tx,a,c,d}*H^{f64,g}_{a}-H^{f64,g}_{a}*K^{f64,g}_{tx,a}+I^{f64,g}_{a}*sqrt(N^{f64,g}_{tx,a}+1.0e-5))/sqrt(N^{f64,g}_{tx,a}+1.0e-5);];$
\\\hline
expand cancellation& it reconstructs the canceled variables to recover the original equation (i.e., the reverse process of cancellation). For example, the equation in the expression $B^{527}_{tx=0}L^{11316}_{a=0}[J^{f64,g}_{tx,a}=1*(C^{f64,g}_{tx,a}*K^{f64,g}_{tx,a}+C^{f64,g}_{tx,a}*M^{f64,g}_{tx,a})/M^{f64,g}_{tx,a};];$ can use the expand cancellation strategy to transform as the expression $B^{527}_{tx=0}L^{11316}_{a=0}[J^{f64,g}_{tx,a}=C^{f64,g}_{tx,a}*(1.0+(K^{f64,g}_{tx,a}/M^{f64,g}_{tx,a}));];$
\\\hline
apart& it decomposes the rational fraction part into simpler partial fractions within one equation (i.e., the reverse process of together). For example, the equation in the expression $B^{339}_{tx=0}L^{1659}_{a=0}[I^{f16,g}_{tx,a}=(K^{f16,g}_{tx,a}+M^{f16,g}_{tx,a})/M^{f16,g}_{tx,a};];$ can use the apart strategy to transform as the expression $B^{339}_{tx=0}L^{1659}_{a=0}[I^{f16,g}_{tx,a}=1.0+(K^{f16,g}_{tx,a}/M^{f16,g}_{tx,a});];$.
\\\hline 
together& it combines multiple fractions into a single fraction within one equation (i.e., the reverse process of apart). For example, the equation in the expression $B^{21}_{tx=0}L^{9036}_{a=0}L^{886}_{c=0}L^{9}_{d=0}[C^{f32,g}_{tx,a,c,d}=D^{f32,g}_{a}*((A^{f32,g}_{tx,a,c,d}-G^{f32,g}_{tx,a})/sqrt(I^{f32,g}_{tx,a}+1e-05))+E^{f32,g}_{a};];$ can use the together strategy to transform as the expression $B^{21}_{tx=0}L^{9036}_{a=0}L^{886}_{c=0}L^{9}_{d=0}[C^{f32,g}_{tx,a,c,d}=(D^{f32,g}_{a}*(A^{f32,g}_{tx,a,c,d}-G^{f32,g}_{tx,a})+E^{f32,g}_{a}*sqrt(I^{f32,g}_{tx,a}+1.0e-5))/sqrt(I^{f32,g}_{tx,a}+1.0e-5);];$. 
\\\hline
power simplification & it simplifies the equation by combining and reducing power operations. For example, the equation in the expression $B^{174}_{tx=0}L^{3662}_{a=0}[D^{f32,g}_{tx,a}=A^{f32,g}_{tx,a}*(1/(1+exp(-A^{f32,g}_{tx,a})));];$ can use the powsimp strategy to transform as the expression $B^{174}_{tx=0}L^{3662}_{a=0}[D^{f32,g}_{tx,a}=A^{f32,g}_{tx,a}/(1+exp(-A^{f32,g}_{tx,a}));];$.\\\hline
expand power simplification& it reverses a simplified power operation into separate power operations within one equation (i.e., the reverse process of power simplification). For example, the equation in the expression $B^{693}_{tx=0}L^{14522}_{a=0}[D^{f16,g}_{tx,a}=A^{f16,g}_{tx,a}/(1+exp(-A^{f16,g}_{tx,a}));];$ can use the expand powsimp strategy to transform as the expression $B^{693}_{tx=0}L^{14522}_{a=0}[D^{f16,g}_{tx,a}=A^{f16,g}_{tx,a}*(1/(1+exp(-A^{f16,g}_{tx,a})));];$.
\\\hline
log simplification& it simplifies logarithmic operations using log properties like product, quotient, and power rules within one equation. For example, the equation in the expression $B^{65}_{tx=0}L^{883}_{a=0}L^{2399}_{c=0}[F^{f16,g}_{tx,a,c}=log(A^{f16,g}_{tx,a,c}*C^{f16,g}_{tx,a,c});];$ can use the logsimp strategy to transform as the expression $B^{65}_{tx=0}L^{883}_{a=0}L^{2399}_{c=0}[F^{f16,g}_{tx,a,c}=log(A^{f16,g}_{tx,a,c})+log(C^{f16,g}_{tx,a,c});];$.\\\hline
expand log simplification & it reverses the simplified logarithm operation into separate log operations within one equation (i.e., the reverse process of log simplification). For example, the equation in the expression $B^{500}_{tx=0}L^{644}_{a=0}L^{3730}_{c=0}[F^{f32,g}_{tx,a,c}=log(A^{f32,g}_{tx,a,c})+log(C^{f32,g}_{tx,a,c});];$ can use the expand log strategy to transform as the expression $B^{500}_{tx=0}L^{644}_{a=0}L^{3730}_{c=0}[F^{f32,g}_{tx,a,c}=log(A^{f32,g}_{tx,a,c}*C^{f32,g}_{tx,a,c});];$.
\\\hline
\end{tabular}
\caption{Mathematical-level atomic optimization strategies 1.}
\label{tab:mathematical-level-strategyv1}
\end{table*}

\begin{table*}[t]
\centering
\begin{tabular}{p{0.3\textwidth} p{0.6\textwidth}} 
\hline
Strategy & Description and Example\\\hline
collect& it combines like operations into a simplified operation in an equation. For example, the equation in the expression $B^{264}_{tx=0}L^{114}_{a=0}[C^{f32,g}_{tx,a}=1.0507010221481323*(max(0,A^{f32,g}_{tx,a})+min(0,1.6732631921768188*(exp(A^{f32,g}_{tx,a})-1)));];$ can use the collect strategy to transform as the expression $B^{264}_{tx=0}L^{114}_{a=0}[C^{f32,g}_{tx,a}=1.0507010221481323*max(0,A^{f32,g}_{tx,a})+1.0507010221481323*min(0,1.6732631921768188*(exp(A^{f32,g}_{tx,a})-1*1));];$.
\\\hline
expand collect& it splits a combined like operation back into multiple operations in an equation (i.e., the reverse process of collect). For example, the equation in the expression $B^{958}_{tx=0}L^{1289}_{a=0}[F^{f32,g}_{tx,a}=0.5*C^{f32,g}_{tx,a}/(1+exp(-C^{f32,g}_{tx,a}));];$ can use the expand collect strategy to transform as the expression $B^{958}_{tx=0}L^{1289}_{a=0}[F^{f32,g}_{tx,a}=(C^{f32,g}_{tx,a}*(1/(1+exp(-C^{f32,g}_{tx,a}))))/2.0;];$.
\\\hline
partially equivalent then correct& it first establishes partial equivalence of several similar expressions by concatenating the input variables, computing the fused expression, and splitting the output variables, as well as finally corrects differences to achieve full equivalence by adding another expression. For example, for the expressions $B^{274}_{tx=0}L^{555}_{a=0}L^{1353}_{c=0}L^{417}_{d=0}L^{3}_{f=0}[D^{f16,g}_{tx,a,c}=D^{f16,g}_{tx,a,c}+H^{f16,g}_{tx,d,c*5+f*6}*G^{f16,g}_{a,d,f};];$ and $B^{274}_{tx=0}L^{555}_{a=0}L^{1353}_{c=0}L^{417}_{d=0}L^{3}_{f=0}[E^{f16,g}_{tx,a,c}=E^{f16,g}_{tx,a,c}+I^{f16,g}_{tx,d,c*5+f*6}*G^{f16,g}_{a,d,f};];$, the inputs can be concatenated by $B^{274}_{tx=0}L^{417}_{a=0}L^{6759}_{c=0}[J^{f16,g}_{tx,a,c}=A^{f16,g}_{tx,a,c};J^{f16,g}_{tx,a,c+6759}=C^{f16,g}_{tx,a,c};];$, the similar operations are executed by $B^{274}_{tx=0}L^{417}_{a=0}L^{13534}_{c=0}[K^{f16,g}_{tx,a,c}=\text{if\_then\_else}(8<=c<13526,J^{f16,g}_{tx,a,c-8},0);];$ and $B^{274}_{tx=0}L^{555}_{a=0}L^{13532}_{c=0}L^{417}_{d=0}L^{3}_{f=0}[M^{f16,g}_{tx,a,c}=M^{f16,g}_{tx,a,c}+K^{f16,g}_{tx,d,c*5+f*6}*G^{f16,g}_{a,d,f};];$, then the output can be split by $B^{274}_{tx=0}L^{555}_{a=0}L^{1353}_{c=0}[D^{f16,g}_{tx,a,c}=M^{f16,g}_{tx,a,c};E^{f16,g}_{tx,a,c}=M^{f16,g}_{tx,a,c+12179};];$, and finally the results are corrected by $B^{274}_{tx=0}L^{555}_{a=0}L^{2}_{c=0}[D^{f16,g}_{tx,a,c+1351}=0;]L^{417}_{d=0}L^{3}_{f=0}[D^{f16,g}_{tx,a,c+1351}=D^{f16,g}_{tx,a,c+1351}+H^{f16,g}_{tx,d,(c+1351)+f*6}*G^{f16,g}_{a,d,f};];B^{274}_{tx=0}L^{555}_{a=0}L^{2}_{c=0}[E^{f16,g}_{tx,a,1-c}=0;]L^{417}_{d=0}L^{3}_{f=0}[E^{f16,g}_{tx,a,1-c}=E^{f16,g}_{tx,a,1-c}+I^{f16,g}_{tx,d,(1-c)+f*6}*G^{f16,g}_{a,d,f};];$
\\\hline
exponential split& it decomposes an exponential operation into multiple factor operations by introducing an existing variable within one equation. For example, for the expression $B^{294}_{tx=0}L^{32193}_{a=0}[C^{f16,g}_{tx,a}=1/(1+exp(-A^{f16,g}_{tx,a}));];$, an exponential term $exp(\text{if\_then\_else}(a-1<0,0,A^{f16,g}_{tx,a-1})$ can be split from the original equation so that the expression $B^{294}_{tx=0}L^{32193}_{a=0}[C^{f16,g}_{tx,a}=1/(1+exp(-A^{f16,g}_{tx,a}-\text{if\_then\_else}(a-1<0,0,A^{f16,g}_{tx,a-1}))*exp(\text{if\_then\_else}(a-1<0,0,A^{f16,g}_{tx,a-1})));];$ is obtained.
\\\hline
multiplicative split& it decomposes one operation into multiple multiplicative factor operations by introducing an existing variable within one equation. For example, for the expression $B^{445}_{tx=0}L^{1}_{a=0}L^{424}_{c=0}L^{207}_{d=0}[F^{f16,g}_{tx,a,c,d}=K^{f16,g}_{tx,a,c,d}/M^{f16,g}_{tx,a,c,d};];$, a multiplicative term $*E^{f16,g}_{tx,a,c,d}$ can be split from the original equation so that the expression $B^{445}_{tx=0}L^{1}_{a=0}L^{424}_{c=0}L^{207}_{d=0}[F^{f16,g}_{tx,a,c,d}=(K^{f16,g}_{tx,a,c,d}/E^{f16,g}_{tx,a,c,d}*E^{f16,g}_{tx,a,c,d})/M^{f16,g}_{tx,a,c,d};];$ is obtained.
\\\hline
additive split& it decomposes one operation into multiple additive factor operations by introducing an existing variable within one equation. For example, for the expression $B^{466}_{tx=0}L^{1}_{a=0}L^{1}_{c=0}L^{2}_{d=0}[J^{f32,g}_{tx,a,c,d}=I^{f32,g}_{tx,a,c,d}+M^{f32,g}_{0,a,c,d};];$, an additive term $-I^{f32,g}_{tx,a,c,d}$ can be split from the original equation so that the expression $B^{466}_{tx=0}L^{1}_{a=0}L^{1}_{c=0}L^{2}_{d=0}[J^{f32,g}_{tx,a,c,d}=I^{f32,g}_{tx,a,c,d}+I^{f32,g}_{tx,a,c,d}-I^{f32,g}_{tx,a,c,d}+M^{f32,g}_{0,a,c,d};];$ is obtained.
\\\hline
\end{tabular}
\caption{Mathematical-level atomic optimization strategies 2.}
\label{tab:mathematical-level-strategyv2}
\end{table*}

\begin{table*}[t]
\centering
\begin{tabular}{p{0.3\textwidth} p{0.6\textwidth}} 
\hline
Strategy & Description and Example\\\hline
normal loop to prefix loop for max operation& it transforms the maximum operations into online streaming operations where the current step is based on the previous step. For example, for the expression $B^{156}_{tx=0}L^{8}_{a=0}L^{274}_{c=0}L^{274}_{d=0}[An^{f16,g}_{tx,a,c}=An^{f16,g}_{tx,a,c}+exp(Y^{f16,g}_{tx,a,c,d}-Am^{f16,g}_{tx,a,c});];$ with the summation operation on the exponential term and its previous expression $B^{156}_{tx=0}L^{8}_{a=0}L^{274}_{c=0}L^{274}_{d=0}[Am^{f16,g}_{tx,a,c}=max(Am^{f16,g}_{tx,a,c},Y^{f16,g}_{tx,a,c,d});];$ with the maximum operation, online streaming method can be used via updating the next step based on the previous step by $B^{156}_{tx=0}L^{8}_{a=0}L^{274}_{c=0}L^{274}_{d=0}[Am^{f16,g}_{tx,a,c,d}=max(\text{if\_then\_else}(d-1<0,-inf,Am^{f16,g}_{tx,a,c,d-1}),Y^{f16,g}_{tx,a,c,d});Ao^{f16,g}_{tx,a,c,d}=\text{if\_then\_else}(d-1<0,1,Ao^{f16,g}_{tx,a,c,d-1})*exp(\text{if\_then\_else}(d-1<0,-inf,Am^{f16,g}_{tx,a,c,d-1})-Am^{f16,g}_{tx,a,c,d})+exp(Y^{f16,g}_{tx,a,c,d}-Am^{f16,g}_{tx,a,c,d});];$ and writing the output by $B^{156}_{tx=0}L^{8}_{a=0}L^{274}_{c=0}[An^{f16,g}_{tx,a,c}=Ao^{f16,g}_{tx,a,c,273};];$.
\\\hline
normal loop to prefix loop for exponential operation& it transforms the summation of exponential operations into online streaming operations where the current step is based on the previous step. Note that this trick uses exponential cancellation to find the prefix relation. For example, for the expression $B^{933}_{tx=0}L^{1}_{a=0}[I^{f16,g}_{tx}=max(I^{f16,g}_{tx},D^{f16,g}_{tx,a});];$ with the maximum operation, online streaming method can be used via initializing the output by $B^{933}_{tx=0}L^{1}_{a=0}[K^{f16,g}_{tx,a}=max(\text{if\_then\_else}(a-1<0,-inf,K^{f16,g}_{tx,a-1}),D^{f16,g}_{tx,a});];$ and then updating the next step based on the previous step by $B^{933}_{tx=0}L^{1}_{a=0}[K^{f16,g}_{tx,a}=max(\text{if\_then\_else}(a-1<0,-inf,K^{f16,g}_{tx,a-1}),D^{f16,g}_{tx,a});];$.
\\\hline
online softmax& it computes the softmax incrementally by updating the tensor variable step by step within a loop nested. For example, for the softmax expressions $B^{186}_{tx=0}L^{1}_{a=0}[I^{f16,g}_{tx}=max(I^{f16,g}_{tx},D^{f16,g}_{tx,a});];B^{186}_{tx=0}L^{1}_{a=0}[J^{f16,g}_{tx}=J^{f16,g}_{tx}+exp(D^{f16,g}_{tx,a}-I^{f16,g}_{tx});];B^{186}_{tx=0}L^{1}_{a=0}[E^{f16,g}_{tx,a}=exp(D^{f16,g}_{tx,a}-I^{f16,g}_{tx})/J^{f16,g}_{tx};];$, the online softmax can be used by$ B^{186}_{tx=0}L^{1}_{a=0}[I^{f16,g}_{tx,a}=max(\text{if\_then\_else}(a-1<0,-inf,I^{f16,g}_{tx,a-1}),D^{f16,g}_{tx,a});K^{f16,g}_{tx,a}=\text{if\_then\_else}(a-1<0,1,K^{f16,g}_{tx,a-1})*exp(\text{if\_then\_else}(a-1<0,-inf,I^{f16,g}_{tx,a-1})-I^{f16,g}_{tx,a})+exp(D^{f16,g}_{tx,a}-I^{f16,g}_{tx,a});];$ and the output can be written by $B^{186}_{tx=0}L^{1}_{a=0}[E^{f16,g}_{tx,a}=exp(D^{f16,g}_{tx,a}-I^{f16,g}_{tx,0})/K^{f16,g}_{tx,0};];$
\\\hline
\end{tabular}
\caption{Mathematical-level atomic optimization strategies 3.}
\label{tab:mathematical-level-strategyv3}
\end{table*}

\begin{table*}[t]
\centering
\begin{tabular}{p{0.3\textwidth} p{0.6\textwidth}} 
\hline
Strategy & Description and Example\\\hline
flashattention without loop tiling& it computes attention expressions in an online manner by incrementally calculating scaled dot-products and applying online softmax. For example, for the softmax expressions $B^{29}_{tx=0}L^{2}_{a=0}L^{3}_{c=0}L^{3}_{d=0}[Ax^{f64,g}_{tx,a,c}=max(Ax^{f64,g}_{tx,a,c},Ac^{f64,g}_{tx,a,c,d});];B^{29}_{tx=0}L^{2}_{a=0}L^{3}_{c=0}L^{3}_{d=0}[Ay^{f64,g}_{tx,a,c}=Ay^{f64,g}_{tx,a,c}+exp(Ac^{f64,g}_{tx,a,c,d}-Ax^{f64,g}_{tx,a,c});];B^{29}_{tx=0}L^{2}_{a=0}L^{3}_{c=0}L^{3}_{d=0}[Ad^{f64,g}_{tx,a,c,d}=exp(Ac^{f64,g}_{tx,a,c,d}-Ax^{f64,g}_{tx,a,c})/Ay^{f64,g}_{tx,a,c};];$ and the matmul expression $B^{29}_{tx=0}L^{2}_{a=0}L^{3}_{c=0}L^{139}_{d=0}L^{3}_{f=0}[Ae^{f64,g}_{tx,a,c,d}=Ae^{f64,g}_{tx,a,c,d}+Ad^{f64,g}_{tx,a,c,f}*X^{f64,g}_{tx,a,f,d};];$, the flashattention without tiling can be used by $B^{29}_{tx=0}L^{2}_{a=0}L^{3}_{c=0}L^{3}_{d=0}[Ax^{f64,l}_{tx,a,c,d}=max(\text{if\_then\_else}(d-1<0,-inf,Ax^{f64,l}_{tx,a,c,d-1}),Ac^{f64,g}_{tx,a,c,d});Az^{f64,l}_{tx,a,c,d}=\text{if\_then\_else}(d-1<0,1,Az^{f64,l}_{tx,a,c,d-1})*exp(\text{if\_then\_else}(d-1<0,-inf,Ax^{f64,l}_{tx,a,c,d-1})-Ax^{f64,l}_{tx,a,c,d})+exp(Ac^{f64,g}_{tx,a,c,d}-Ax^{f64,l}_{tx,a,c,d});L^{139}_{i=0}[Ca^{f64,g}_{tx,a,c,i,d}=\text{if\_then\_else}(d-1<0,1,Ca^{f64,g}_{tx,a,c,i,d-1})*\text{if\_then\_else}(d-1<0,1,Az^{f64,l}_{tx,a,c,d-1})*exp(\text{if\_then\_else}(d-1<0,-inf,Ax^{f64,l}_{tx,a,c,d-1})-Ax^{f64,l}_{tx,a,c,d})/Az^{f64,l}_{tx,a,c,d}+exp(Ac^{f64,g}_{tx,a,c,d}-Ax^{f64,l}_{tx,a,c,d})/Az^{f64,l}_{tx,a,c,d}*X^{f64,g}_{tx,a,d,i};];];$ and the output can be written by $B^{29}_{tx=0}L^{139}_{d=0}L^{2}_{a=0}L^{3}_{c=0}[Ae^{f64,g}_{tx,a,c,d}=Ca^{f64,g}_{tx,a,c,d,2};];$.
\\\hline
normal loop to prefix loop for matmul operation based on online softmax& it transforms a standard matrix multiplication into an online prefix computation based on online softmax applied in previous computations. For example, for the online softmax expressions $B^{110}_{tx=0}L^{8}_{a=0}L^{2}_{c=0}L^{2}_{d=0}[As^{f32,g}_{tx,a,c,d}=max(\text{if\_then\_else}(d-1<0,-inf,As^{f32,g}_{tx,a,c,d-1}),Y^{f32,g}_{tx,a,c,d});Cc^{f32,g}_{tx,a,c,d}=\text{if\_then\_else}(d-1<0,1,Cc^{f32,g}_{tx,a,c,d-1})*exp(\text{if\_then\_else}(d-1<0,-inf,As^{f32,g}_{tx,a,c,d-1})-As^{f32,g}_{tx,a,c,d})+exp(Y^{f32,g}_{tx,a,c,d}-As^{f32,g}_{tx,a,c,d});];B^{110}_{tx=0}L^{8}_{a=0}L^{2}_{c=0}L^{2}_{d=0}[Z^{f32,g}_{tx,a,c,d}=exp(Y^{f32,g}_{tx,a,c,d}-As^{f32,g}_{tx,a,c,1})/Cc^{f32,g}_{tx,a,c,1};];$ and the matmul expression $B^{110}_{tx=0}L^{8}_{a=0}L^{2}_{c=0}L^{50}_{d=0}L^{2}_{f=0}[Aa^{f32,g}_{tx,a,c,d}=Aa^{f32,g}_{tx,a,c,d}+Z^{f32,g}_{tx,a,c,f}*N^{f32,g}_{tx,a,f,d};];$, the online method can be used by $B^{110}_{tx=0}L^{8}_{a=0}L^{2}_{c=0}L^{2}_{d=0}[As^{f32,l}_{tx,a,c,d}=max(\text{if\_then\_else}(d-1<0,-inf,As^{f32,l}_{tx,a,c,d-1}),Y^{f32,g}_{tx,a,c,d});Cc^{f32,l}_{tx,a,c,d}=\text{if\_then\_else}(d-1<0,1,Cc^{f32,l}_{tx,a,c,d-1})*exp(\text{if\_then\_else}(d-1<0,-inf,As^{f32,l}_{tx,a,c,d-1})-As^{f32,l}_{tx,a,c,d})+exp(Y^{f32,g}_{tx,a,c,d}-As^{f32,l}_{tx,a,c,d});L^{50}_{i=0}[Cd^{f32,g}_{tx,a,c,i,d}=\text{if\_then\_else}(d-1<0,1,Cd^{f32,g}_{tx,a,c,i,d-1})*\text{if\_then\_else}(d-1<0,1,Cc^{f32,l}_{tx,a,c,d-1})*exp(\text{if\_then\_else}(d-1<0,-inf,As^{f32,l}_{tx,a,c,d-1})-As^{f32,l}_{tx,a,c,d})/Cc^{f32,l}_{tx,a,c,d}+exp(Y^{f32,g}_{tx,a,c,d}-As^{f32,l}_{tx,a,c,d})/Cc^{f32,l}_{tx,a,c,d}*N^{f32,g}_{tx,a,d,i};];];$ and the output can be written by $B^{110}_{tx=0}L^{50}_{d=0}L^{2}_{c=0}L^{8}_{a=0}[Aa^{f32,g}_{tx,a,c,d}=Cd^{f32,g}_{tx,a,c,d,1};];$.
\\
\hline
\end{tabular}
\caption{Mathematical-level atomic optimization strategies 4.}
\label{tab:mathematical-level-strategyv4}
\end{table*}

\textbf{Scope of Tensor-Program-Level Optimization.} Tensor-program-level optimization focuses on the logical transformations of a program's structure and algebraic representation, specifically targeting loop hierarchies, equation compositions, and tensor access patterns without altering computational semantics. While kernel- or hardware-level tuning manages physical execution details (e.g., memory bank conflicts, cache line utilization, and pipeline depths), this logical abstraction defines what is computed and how it is logically structured, independent of its implementation on specific hardware targets.

\subsection{Strategy Categories}

\textbf{Motivation for Categorizing Strategies.} We categorize tensor-program-level optimizations into distinct strategies for two main reasons: 
(i). Facilitating step-level supervision. Effective supervision requires models to identify the semantic shift introduced at each transformation step. However, due to the vast tensor-program-level optimization space, each step often induces complex and entangled program changes, leading to supervision signals that are parse and semantically coarse. To ensure learnability, we partition the strategy space into a structured taxonomy.
(ii). Handling strategy heterogeneity. Beyond mere partitioning, the optimization space benefits from organization into distinct semantic levels due to its inherent heterogeneity. Transformations at the tensor-program level act on diverse scopes (e.g., loop structure, algebraic formation), each requiring fundamentally different reasoning logic. Treating these as a flat space would obscure such structural differences, complicating the learning of disparate optimization logics. We therefore organize the strategy space by the program aspect being modified, ensuring each level represents a consistent semantic domain. 
Consequently, this stratification further allows complex end-to-end optimizations to be decomposed into interpretable and analyzable sequences of strategy steps.

\textbf{Four Categories of Strategies.} Based on the semantic dimension primarily modified by a strategy, we categorize tensor-program-level optimizations into four classes, as detailed in Table~\ref{tab:graph-level-strategy}-Table~\ref{tab:mathematical-level-strategyv4}

(1). Graph level. These optimizations act on the organization of multiple expressions. An example is the expression reorder strategy, where independent expressions are permuted without violating data dependencies.

(2). Operator level. These strategies target the loop–equation structure within a single expression. For example, the loop reorder strategy swaps the nesting order of LEIR iterators, such as transforming $B^{719}_{tx=0}L^{549}_{a=0}$ into $L^{549}_{a=0}B^{719}_{tx=0}$.

(3). Memory level. These optimizations primarily alter the logical storage and layouts of tensor variables. the storage scope strategy rebinds a tensor $E^{f64,g}_{a}$ in global memory to a local memory scope $E^{f64,l}_{a}$ to optimize data proximity.

(4). Mathematical level. These strategies directly rewrite algebraic formulations to change execution logic while preserving numerical semantics. For example, the online softmax strategy reformulates the global exponential sum into incremental update equations, decomposing a monolithic reduction into recursive algebraic steps within a loop.

In total, we have identified 43 distinct strategies across these levels, comprising 8 graph-level, 9 operator-level, 5 memory-level, and 21 mathematical-level strategies. Among these, mathematical-level strategies are particularly challenging to realize with modular or rule-based tools as they require global algebraic reasoning. LEIR addresses these challenges by providing a high-density, unified representation that abstracts these disparate dimensions into a standardized, step-level format. 

\subsection{Preconditions for Strategy Filtering}
Given that strategies are constrained by specific program structures, we define nine essential preconditions as follows: (1). The pattern match check detects a specific computation pattern (e.g., softmax computation for online softmax strategy); (2). The dependency check verifies the existence of computation-order dependencies among expressions (e.g., for expression reorder strategy); (3). The operation identity check ensures that repeated computation operations exist (e.g., for common subexpression elimination); (4). The loop nest consistency check ensures that expressions share the same loop structure (e.g., for operator fusion strategy); (5). The equation count check ensures a sufficient number of equations exist (e.g., for operator fission strategy); (6). The loop axis count check verifies that the number of loop axes is sufficient (e.g., for loop reorder strategy); (7). The loop range factorization check ensures that the selected loop range can be split appropriately (e.g., for loop split strategy); (8). The reduction axis check prevents applying illegal strategies (e.g., loop binding) to reduction axes; (9). The intermediate variable check validates whether the tensor variables are inputs or outputs (e.g., for set storage scope strategy).

\begin{enumerate}
    \item operator fusion: dependency check, loop nest consistency check;
\item operator fission: dependency check, equation count check;
\item compute inline: dependency check, loop nest consistency check;
\item expression splitting: n/a;
\item tensor concat to fuse operators: dependency check, operation identity check;
\item tensor split to decouple operators: n/a;
\item common subexpression elimination: operation identity check, loop nest consistency check;
\item expression reorder: dependency check ;
\item loop reorder: loop axis count check;
\item loop tiling: loop axis count check, loop range factorization check;
\item loop split: loop range factorization check;
\item loop fusion: loop axis count check, reduction axis check;
\item loop unrolling: n/a;
\item loop parallelization: reduction axis check;
\item loop vectorization: reduction axis check;
\item loop binding: reduction axis check;
\item reduction factorization: reduction axis check;
\item cache read write: n/a;
\item layout transformation: n/a;
\item set storage scope: intermediate variable check;
\item set storage layout: intermediate variable check;
\item precompute indices: pattern match check;
\item factorization: pattern match check;
\item expand factorization: pattern match check;
\item cancellation: pattern match check;
\item expand cancellation: pattern match check;
\item apart: pattern match check;
\item together: pattern match check;
\item powsimp: pattern match check;
\item expand powsimp: pattern match check;
\item logsimp: pattern match check;
\item expand log: pattern match check;
\item collect: pattern match check;
\item expand collect: pattern match check;
\item partially equivalent then correct: n/a;
\item exponential split: pattern match check;
\item multiplicative split: n/a;
\item additive split: n/a;
\item normal loop max to prefix max: pattern match check;
\item normal loop summation on exp to prefix summation on exp: pattern match check;
\item online softmax: pattern match check;
\item flashattention wo tiling: pattern match check  
\item normal matmul to prefix matmul based on online softmax: pattern match check.
\end{enumerate}

\subsection{Strategy Difficulty}
\begin{itemize}
    \item Easy: operator fission, factorization, expand factorization, cancellation, expand cancellation, apart, together, powsimp, expand powsimp, logsimp, expand log, collect, expand collect;

    \item Medium: operator fusion, compute inline, expression splitting, expression reorder, loop reorder, loop unrolling, loop parallelization, loop vectorization, loop binding, exponential split, ultiplicative split, additive split;

    \item Dfficult:tensor concat to fuse operators, tensor split to decouple operators, common subexpression elimination, loop tiling, loop split, loop fusion, reduction factorization, cache read write, layout transformation, set storage scope
, set storage layout, precompute indices, partially equivalent then correct, normal loop max to prefix max, normal loop summation on exp to prefix summation on exp, online softmax, flashattention wo tiling, normal matmul to prefix matmul based on online softmax; 

\end{itemize}

\section{Evaluation}\label{sec:appendix-eval}

\textbf{Training.} All models are fine-tuned using LoRA with a rank of $8$, LoRA alpha set to $32$, and a dropout rate of $0.05$. LoRA adapters are applied to all linear layers. We use a learning rate of $1e-4$, a weight decay of $0.1$, and a warmup ratio of $0.05$.
The batch size is set to $64$ for Qwen3-1.7B and $16$ for the larger models due to memory constraints. All models are trained for 3 epochs. LoRA fine-tuning is implemented using the ms-swift framework.

\subsection{Single-step Transformation}
\textbf{Setup.} We evaluate all models trained on our dataset across 2248 test cases, constructed from 180 distinct tensor programs by varying input/output data types and tensor shapes. All test cases are strictly held out from training. The workload of these test cases are detailed at the end of the section. 
Each model is tasked to generate a mathematically equivalent IR in a single step, following the prompt format of our dataset's single-answer version. The temperature is set to 0 during inference. Crucially, we do not specify any strategies, requiring the model to autonomously select a valid transformation.

\subsection{Multi-step Transformation}
\textbf{Setup.} We evaluate Qwen3-32B, trained on our dataset, across 100 distinct tensor programs strictly held out from training. The temperature is set to 0.3 during inference. For accurate performance measurement, we employ TVM's built-in \texttt{time\_evaluator} to record execution time. Specifically, each kernel is executed for three warmup runs to mitigate transient hardware effects, followed by three measurement repetitions to ensure statistical stability, with the average latency reported.

The model is tasked with generating mathematically equivalent, runtime-performance-optimized IRs to achieve higher speedups. Each test case prompt provides six key pieces of information: (i) the search algorithm, (ii) the current LEIR and related metadata, (iii) the exploration history if required by the search algorithm, with at most one parent node, (iv) the target GPU specifications, (v) all 43 potential strategies, and (vi) the task description. An example prompt is provided as follows:

"Breadth-first-based optimization is used on a given IR to improve performance. Each IR is a state, and has a parent transformation and speedup performance.

Give the current IR of Gemm Swish Divide Clamp Tanh Clamp: $B^{728}_{tx=0}B^{1243}_{bxa=0}L^{2022}_{c=0}[C^{f32,g}_{tx,bxa}=C^{f32,g}_{tx,bxa}+A^{f32,g}_{tx,c}*J^{f32,g}_{bxa,c};];B^{728}_{tx=0}L^{1243}_{a=0}[C^{f32,g}_{tx,a}$ $=C^{f32,g}_{tx,a}+K^{f32,g}_{a};];B^{728}_{tx=0}L^{1243}_{a=0}[F^{f32,g}_{tx,a}=0.5*C^{f32,g}_{tx,a}/(1+exp(-C^{f32,g}_{tx,a}$ $));];$ $B^{728}_{tx=0}L^{1243}_{a=0}[G^{f32,g}_{tx,a}=min(max(F^{f32,g}_{tx,a}$ $,-1.0),1.0);];B^{728}_{tx=0}L^{1243}_{a=0}$ $[M^{f32,g}_{tx,a}=exp(G^{f32,g}_{tx,a})-exp(-G^{f32,g}_{tx,a});];$ $B^{728}_{tx=0}L^{1243}_{a=0}[N^{f32,g}_{tx,a}=exp$ $(G^{f32,g}_{tx,a})+exp(-G^{f32,g}_{tx,a});];B^{728}_{tx=0}L^{1243}_{a=0}[H^{f32,g}_{tx,a}$ $=M^{f32,g}_{tx,a}/N^{f32,g}_{tx,a};];$ $B^{728}_{tx=0}$ $L^{1243}_{a=0}[I^{f32,g}_{tx,a}=min(max(H^{f32,g}_{tx,a},-1.0),1.0);];$, where the known variables are 'A' with the dtype torch.float32 and shape [728, 2022], 'J' with the dtype torch.float32 and shape [1243, 2022], 'K' with the dtype torch.float32 and shape [1243], and 'I' with the dtype torch.float32 and shape [728, 1243]. Do not change the names, shapes or dtypes of these known variables in the IR. 

History:

The parent IR: same as the root IR, depth:0, speedup value: 1. 

The speedup value of the current IR: 30.358607118641192, the depth is 1, and the current IR is obtained from the parent IR using the strategy 'loop\_binding'.

**Target hardware**: NVIDIA H20-3e GPU.

CUDA binding rules: Loop axes bound to block (along x,y,z axis, max dimension value: $2^31-1$, 65535, 65535) MUST be renamed with prefixes bx, by, bz and unique, respectively, followed by other unique lowercase letters. Loop axes bound to thread (along x,y,z axis, max dimension value: 1024, 1024, 64) MUST be renamed with prefixes tx, ty, tz, respectively, followed by other unique lowercase letters.

Memory usage rules: Data indexed by block-level loops may be placed in shared (s) or global (g) memory. Data indexed by thread-level loops may be placed in local (l), shared (s) or global (g) memory.

The following strategies and any other mathematical strategies can be considered:
operator fusion, operator fission, compute inline, expression splitting, tensor concat to fuse operators, tensor split to decouple operators, common subexpression elimination, expression reorder, loop reorder, loop tiling, loop split, loop fusion, loop unrolling, loop parallelization, loop vectorization, loop binding, reduction factorization, cache read write, layout transformation, set storage scope, set storage layout, precompute indices, factorization, expand factorization, cancellation,expand cancellation, apart, together, powsimp, expand powsimp, expand log, logsimp, collect, expand collect, partially equivalent then correct, normal loop max to prefix max, exponential split, multiplicative split, additive split, normal loop summation on exp to prefix summation on exp, online softmax, flashattention wo tiling, normal matmul to prefix matmul based on online softmax.

**Task**:

Please give me at least 2 different **numerically equivalent, runtime-performance-optimized** IRs that produce exactly the same outputs for any floating-point inputs (bitwise identical) to achieve higher speedup values (should be more than 1), and also provide applied strategy for each transformed IR. 

Return the answer list **only** as a valid JSON object, and each entry with the following keys: 'idx', 'transformed\_IR', 'applied\_strategies'.

CRITICAL:

1. Before you suggest each new transformation, you MUST identify what has been changed in the current IR compared to the root IR. For each new optimization, you MUST build it ON TOP OF these existing changes, namely ON TOP OF the current IR. The strategies MUST be used on the current IR! You MUST compare your modified parts in each transformed IR with the current IR. If they are identical strings, your answer is WRONG. If the unmodified part in each transformed IR is different from the current IR, your answer is WRONG.

2.Don't repeat the current or parent IRs! You MUST NOT revert to the parent IR: In particular, you are NOT allowed to apply any reverse or undo operation that reconstructs the current IR from its parent IR, including inverse transformations such as operator fusion <-> operator fission, loop tiling <-> loop fusion, loop split <-> loop fusion, apart <-> together, collect <-> expand collect, or similar reversals."

\textbf{Search Algorithm.} To implement multi-step optimization, we deploy seven distinct search algorithms categorized by their exploration strategies: (i) Greedy Search, which generates multiple candidate transformations per step and selects the locally best one; (ii) Breadth-First Search (BFS) and (iii) Depth-First Search (DFS), representing exhaustive breadth and depth explorations; (iv) Beam Search, which generates multiple transformations per step and maintains the top-$k$ candidates as the search frontier; (v) Monte Carlo Tree Search (MCTS), which balances exploration and exploitation using rollouts and value estimation to guide transformation selection; and (vi) Chain-based Search (with/without parent nodes), which sequentially refines transformations step by step, optionally considering the parent node. 

For greedy search, beam search, MCTS, and two chain-based search variants, the maximum number of iterations is set to 20. For DFS and BFS, the maximum depth is set to 20 andthe number of generated LEIRs for each node is set to 2. For greedy search, the breadth is set to 2. For beam search, the number of generated LEIRs for each node is set to 3, and the k value is set to 2 in order to maintain the top-k candidates.

\textbf{Case Study.} We analyze a complex tensor program featuring matrix multiplication, scaling, and residual addition(i.e., $B^{457}_{tx=0}L^{2265}_{a=0}L^{3520}_{c=0}$ $[D^{f32,g}_{tx,a}=D^{f32,g}_{tx,a}+A^{f32,g}_{tx,c}*I^{f32,g}_{a,c};];B^{457}_{tx=0}L^{2265}_{a=0}[D^{f32,g}_{tx,a}=D^{f32,g}_{tx,a}+J^{f32,g}_{a};];B^{457}_{tx=0}L^{2265}_{a=0}[E^{f32,g}_{tx,a}=D^{f32,g}_{tx,a}*H^{f32,g}_{a};];B^{457}_{tx=0}L^{2265}_{a=0}[F^{f32,g}_{tx,a}=E^{f32,g}_{tx,a}+C^{f32,g}_{tx,a};];$), achieving a $561.82\times$ speedup through a 15-step optimization trajectory. The process begins with a layout transposition (Step 1, $8.34\times$) to align memory access. This is followed by intensive memory hierarchy and parallelism adjustments—including storage scope specification and loop binding—to reach $288.75\times$ (Steps 2–6). Subsequent multi-level tiling and unrolling further optimize data locality and register pressure to $430.68\times$ (Steps 7–12), concluding with fine-grained hardware binding to maximize GPU utilization (Steps 13–15). The final optimized LEIR is: $B^{457}_{tx=0}L^{3520}_{c=0}[M^{f32,l}_{c,tx}=A^{f32,g}_{tx,c};];B^{457}_{tx=0}L^{3520}_{c=0}[K^{f32,g}_{c,tx}=M^{f32,l}_{c,tx};];B^{457}_{tx=0}U^{3520}_{c=0}[N^{f32,g}_{c*457+tx}=K^{f32,g}_{c,tx}$ $;];B^{457}_{tx=0}B^{2265}_{bxa=0}U^{3520}_{c=0}[D^{f32,s}_{tx,bxa}=D^{f32,s}_{tx,bxa}+N^{f32,g}_{c*457+tx}*I^{f32,g}_{bxa,c};];B^{457}_{tx=0}$ $B^{2265}_{bza=0}[D^{f32,s}_{tx,bza}=D^{f32,s}_{tx,bza}+J^{f32,g}_{bza};];B^{457}_{tx=0}B^{2265}_{bza=0}[E^{f32,g}_{tx,bza}=D^{f32,s}_{tx,bza}*H^{f32,g}_{bza};];B^{457}_{tx=0}B^{5}_{bxf=0}L^{453}_{a=0}[F^{f32,g}_{tx,bxf*453+a}=E^{f32,g}_{tx,bxf*453+a}+C^{f32,g}_{tx,bxf*453+a}$ $;];$.

\textbf{Workload of testcases.} 
\begin{enumerate}
    \item InstanceNorm: the known variables are 'A' with the dtype torch.float16 and shape [67, 27, 77, 10], 'D' with the dtype torch.float16 and shape [27], 'E' with the dtype torch.float16 and shape [27], and 'C' with the dtype torch.float16 and shape [67, 27, 77, 10]. 

    \item Square matrix multiplication: the known variables are 'A' with the dtype torch.float32 and shape [319, 319], 'C' with the dtype torch.float32 and shape [319, 319], and 'D' with the dtype torch.float32 and shape [319, 319]. 

     \item conv depthwise 2D square input square kernel:  the known variables are 'A' with the dtype torch.float64 and shape [64, 63, 49, 366], 'D' with the dtype torch.float64 and shape [63, 1, 2, 2], and 'C' with the dtype torch.float64 and shape [64, 63, 25, 184]. 

    \item Mean reduction over a dimension: the known variables are 'A' with the dtype torch.float16 and shape [16, 256, 256], and 'C' with the dtype torch.float16 and shape [16, 256]. 

     \item Gemm Sigmoid Sum LogSumExp: the known variables are 'A' with the dtype torch.float32 and shape [128, 10], 'G' with the dtype torch.float32 and shape [20, 10], 'H' with the dtype torch.float32 and shape [20], and 'F' with the dtype torch.float32 and shape []. 

    \item Product reduction over a dimension:  the known variables are 'A' with the dtype torch.float32 and shape [301, 3630, 744], and 'C' with the dtype torch.float32 and shape [301, 744]. 

\item Conv3d HardSwish ReLU Softmax Mean: the known variables are 'A' with the dtype torch.float16 and shape [120, 2, 8, 22, 56], 'H' with the dtype torch.float16 and shape [52, 2, 4, 4, 4], and 'G' with the dtype torch.float16 and shape [120, 52]. 

\item Matmul with small K dimension: the known variables are 'A' with the dtype torch.float16 and shape [1024, 1024], 'C' with the dtype torch.float16 and shape [1024, 32], and 'D' with the dtype torch.float16 and shape [1024, 32]. 

 \item LeakyReLU: the known variables are 'A' with the dtype torch.float16 and shape [931, 3862], and 'C' with the dtype torch.float16 and shape [931, 3862]. 

 \item Matmul with transposed both:  the known variables are 'A' with the dtype torch.float64 and shape [120, 543], 'C' with the dtype torch.float64 and shape [808, 120], and 'F' with the dtype torch.float64 and shape [543, 808]. 

     \item Matmul Subtract Multiply ReLU: the known variables are 'A' with the dtype torch.float16 and shape [738, 715], 'G' with the dtype torch.float16 and shape [], 'H' with the dtype torch.float16 and shape [], 'I' with the dtype torch.float16 and shape [3251, 715], 'J' with the dtype torch.float16 and shape [3251], and 'F' with the dtype torch.float16 and shape [738, 3251]. 

     \item Gemm BiasAdd Hardtanh Mish GroupNorm: the known variables are 'A' with the dtype torch.float32 and shape [924, 1220], 'H' with the dtype torch.float32 and shape [16], 'I' with the dtype torch.float32 and shape [16], 'J' with the dtype torch.float32 and shape [16, 1220], 'K' with the dtype torch.float32 and shape [16], 'M' with the dtype torch.float32 and shape [16], 'N' with the dtype torch.float32 and shape [16], and 'G' with the dtype torch.float32 and shape [924, 16]. 

    \item Conv3d Softmax MaxPool MaxPool: the known variables are 'A' with the dtype torch.float32 and shape [69, 38, 38, 6, 45], 'G' with the dtype torch.float32 and shape [5, 38, 2, 2, 2], and 'F' with the dtype torch.float32 and shape [69, 5, 37, 5, 44]. 

     \item GQA sum: the known variables are 'A' with the dtype torch.float32 and shape [715, 105, 140], 'Ak' with the dtype torch.bool and shape [1, 1, 105, 105], 'Am' with the dtype torch.float32 and shape [140, 140], 'An' with the dtype torch.float32 and shape [140], 'Ao' with the dtype torch.float32 and shape [35, 140], 'Aq' with the dtype torch.float32 and shape [35], 'Ar' with the dtype torch.float32 and shape [35, 140], 'As' with the dtype torch.float32 and shape [35], 'Aw' with the dtype torch.float32 and shape [140, 140], 'Ax' with the dtype torch.float32 and shape [140], and 'Aj' with the dtype torch.float32 and shape [715, 1, 140]. 

     \item Matmul Mean Softmax: the known variables are 'A' with the dtype torch.float32 and shape [128, 100], 'F' with the dtype torch.float32 and shape [50, 100], 'G' with the dtype torch.float32 and shape [50], and 'E' with the dtype torch.float32 and shape [128, 1]. 

    \item Conv3d GroupNorm Mean: the known variables are 'A' with the dtype torch.float16 and shape [115, 6, 9, 6, 48], 'F' with the dtype torch.float16 and shape [16, 6, 1, 1, 1], 'G' with the dtype torch.float16 and shape [16], 'H' with the dtype torch.float16 and shape [16], and 'E' with the dtype torch.float16 and shape [115]. 
    
    \item  L1Norm: the known variables are 'A' with the dtype torch.float32 and shape [24, 20767], and 'E' with the dtype torch.float32 and shape [24, 20767]. 

   \item   Gemm Sigmoid Scaling ResidualAdd: the known variables are 'A' with the dtype torch.float16 and shape [275, 776], 'G' with the dtype torch.float16 and shape [], 'H' with the dtype torch.float16 and shape [776, 776], 'I' with the dtype torch.float16 and shape [776], and 'F' with the dtype torch.float16 and shape [275, 776]. 

   \item   Matmul Min Subtract: the known variables are 'A' with the dtype torch.float64 and shape [903, 763], 'F' with the dtype torch.float64 and shape [], 'G' with the dtype torch.float64 and shape [], 'H' with the dtype torch.float64 and shape [2249, 763], 'I' with the dtype torch.float64 and shape [2249], and 'E' with the dtype torch.float64 and shape [903, 2249]. 

    \item Matmul Swish Scaling:  the known variables are 'A' with the dtype torch.float64 and shape [647, 1930], 'G' with the dtype torch.float64 and shape [], 'H' with the dtype torch.float64 and shape [2899, 1930], 'I' with the dtype torch.float64 and shape [2899], and 'F' with the dtype torch.float64 and shape [647, 2899]. 

    \item  Gemm Swish Divide Clamp Tanh Clamp:  the known variables are 'A' with the dtype torch.float32 and shape [728, 2022], 'J' with the dtype torch.float32 and shape [1243, 2022], 'K' with the dtype torch.float32 and shape [1243], and 'I' with the dtype torch.float32 and shape [728, 1243]. 

\item  CrossEntropyLoss:  the known variables are 'A' with the dtype torch.float16 and shape [603, 777], 'C' with the dtype torch.int64 and shape [603], and 'D' with the dtype torch.float16 and shape []. 

 \item    BMM InstanceNorm Sum ResidualAdd Multiply:  the known variables are 'A' with the dtype torch.float64 and shape [270, 1026], 'C' with the dtype torch.float64 and shape [270, 497, 2], 'M' with the dtype torch.float64 and shape [497, 1026], 'N' with the dtype torch.float64 and shape [497], 'O' with the dtype torch.float64 and shape [497], 'Q' with the dtype torch.float64 and shape [497], and 'K' with the dtype torch.float64 and shape [270, 497, 2]. 

  \item    Conv2d GroupNorm Scale MaxPool Clamp:  the known variables are 'A' with the dtype torch.float64 and shape [128, 3, 32, 32], 'H' with the dtype torch.float64 and shape [16, 1, 1], 'I' with the dtype torch.float64 and shape [16, 1, 1], 'J' with the dtype torch.float64 and shape [16, 3, 3, 3], 'K' with the dtype torch.float64 and shape [16], 'M' with the dtype torch.float64 and shape [16], and 'G' with the dtype torch.float64 and shape [128, 16, 15, 15]. 

  \item   MHA max: the known variables are 'A' with the dtype torch.float64 and shape [622, 5, 498], 'Ai' with the dtype torch.bool and shape [1, 1, 5, 5], 'Aj' with the dtype torch.float64 and shape [1494, 498], 'Ak' with the dtype torch.float64 and shape [1494], 'Am' with the dtype torch.float64 and shape [498, 498], 'An' with the dtype torch.float64 and shape [498], and 'Ah' with the dtype torch.float64 and shape [622, 1, 498]. 
    
 \item LogSoftmax the known variables are 'A' with the dtype torch.float32 and shape [304, 12782], and 'C' with the dtype torch.float32 and shape [304, 12782]. 
    
  \item  LogSoftmax: the known variables are 'A' with the dtype torch.float32 and shape [304, 12782], and 'C' with the dtype torch.float32 and shape [304, 12782]. 

    \item  Matrix scalar multiplication: the known variables are 'A' with the dtype torch.float64 and shape [778, 2385], 'C' with the dtype torch.float64 and shape [], and 'D' with the dtype torch.float64 and shape [778, 2385].

  \item   MSELoss: the known variables are 'A' with the dtype torch.float16 and shape [426, 64], 'C' with the dtype torch.float16 and shape [426, 64], and 'F' with the dtype torch.float16 and shape []. 

   \item  Swish: the known variables are 'A' with the dtype torch.float16 and shape [32, 11421], and 'D' with the dtype torch.float16 and shape [32, 11421]. 

   \item   Gemm Subtract GlobalAvgPool LogSumExp GELU ResidualAdd: the known variables are 'A' with the dtype torch.float64 and shape [10, 564], 'I' with the dtype torch.float64 and shape [1313], 'J' with the dtype torch.float64 and shape [1313], 'K' with the dtype torch.float64 and shape [1313, 564], 'M' with the dtype torch.float64 and shape [1313], and 'H' with the dtype torch.float64 and shape [10, 564]. 

  \item   Average Pooling 1D:  the known variables are 'A' with the dtype torch.float16 and shape [787, 1097, 481], and 'C' with the dtype torch.float16 and shape [787, 1097, 475]. 

  \item    HardSigmoid:  the known variables are 'A' with the dtype torch.float64 and shape [129, 7733], and 'C' with the dtype torch.float64 and shape [129, 7733]. 

    \item  Gemm Multiply LeakyReLU: the known variables are 'A' with the dtype torch.float16 and shape [654, 1472], 'F' with the dtype torch.float16 and shape [], 'G' with the dtype torch.float16 and shape [738, 1472], 'H' with the dtype torch.float16 and shape [738], and 'E' with the dtype torch.float16 and shape [654, 738]. 

   \item   MHA Gemm ReLU sum:  the known variables are 'A' with the dtype torch.float64 and shape [264, 37, 92], 'Aj' with the dtype torch.bool and shape [1, 1, 37, 37], 'Ak' with the dtype torch.float64 and shape [276, 92], 'Am' with the dtype torch.float64 and shape [276], 'An' with the dtype torch.float64 and shape [92, 92], 'Ao' with the dtype torch.float64 and shape [92], 'Aq' with the dtype torch.float64 and shape [935, 92], and 'Ai' with the dtype torch.float64 and shape [264, 1, 935]. 

 \item    Max Pooling 1D: the known variables are 'A' with the dtype torch.float16 and shape [16, 64, 128], and 'C' with the dtype torch.float16 and shape [16, 64, 62]. 

  \item    Average Pooling 2D:  the known variables are 'A' with the dtype torch.float16 and shape [19, 160, 945, 5], and 'C' with the dtype torch.float16 and shape [19, 160, 236, 1]. 

   \item   Max Pooling 2D:  the known variables are 'A' with the dtype torch.float16 and shape [16, 32, 128, 128], and 'C' with the dtype torch.float16 and shape [16, 32, 64, 64]. 

  \item   Conv3d GroupNorm Min Clamp:  the known variables are 'A' with the dtype torch.float32 and shape [79, 47, 59, 35, 5], 'G' with the dtype torch.float32 and shape [], 'H' with the dtype torch.float32 and shape [8, 47, 3, 3, 3], 'I' with the dtype torch.float32 and shape [8], 'J' with the dtype torch.float32 and shape [8], and 'F' with the dtype torch.float32 and shape [79, 8, 57, 33, 3]. 

  \item   Matmul MaxPool Sum Scale:  the known variables are 'A' with the dtype torch.float32 and shape [200, 3752], 'I' with the dtype torch.float32 and shape [], 'J' with the dtype torch.float32 and shape [2588, 3752], 'K' with the dtype torch.float32 and shape [2588], and 'H' with the dtype torch.float32 and shape [200]. 

  \item   MinGPTNewGelu:  the known variables are 'A' with the dtype torch.float32 and shape [568, 11216], and 'J' with the dtype torch.float32 and shape [568, 11216]. 

   \item  FrobeniusNorm: the known variables are 'A' with the dtype torch.float16 and shape [106, 29, 29, 89], and 'D' with the dtype torch.float16 and shape [106, 29, 29, 89]. 

  \item   TripletMarginLoss:  the known variables are 'A' with the dtype torch.float16 and shape [419, 1834], 'C' with the dtype torch.float16 and shape [419, 1834], 'D' with the dtype torch.float16 and shape [419, 1834], and 'E' with the dtype torch.float16 and shape []. 

   \item  MQA mean:  the known variables are 'A' with the dtype torch.float64 and shape [101, 63, 376], 'Af' with the dtype torch.bool and shape [1, 1, 63, 63], 'Ag' with the dtype torch.float64 and shape [376, 376], 'Ah' with the dtype torch.float64 and shape [376], 'Ai' with the dtype torch.float64 and shape [47, 376], 'Aj' with the dtype torch.float64 and shape [47], 'Ak' with the dtype torch.float64 and shape [47, 376], 'Am' with the dtype torch.float64 and shape [47], 'An' with the dtype torch.float64 and shape [376, 376], 'Ao' with the dtype torch.float64 and shape [376], and 'Ae' with the dtype torch.float64 and shape [101, 1, 376]. 

  \item   Tanh:  the known variables are 'A' with the dtype torch.float32 and shape [776, 5482], and 'C' with the dtype torch.float32 and shape [776, 5482]. 

   \item  Softsign:  the known variables are 'A' with the dtype torch.float32 and shape [927, 11836], and 'E' with the dtype torch.float32 and shape [927, 11836]. 

    \item Matrix Multiplication: the known variables are 'A' with the dtype torch.float16 and shape [550, 54], 'C' with the dtype torch.float16 and shape [54, 550], and 'D' with the dtype torch.float16 and shape [550, 550]. 

    \item Matmul Mish Mish:  the known variables are 'A' with the dtype torch.float32 and shape [775, 177], 'F' with the dtype torch.float32 and shape [2435, 177], 'G' with the dtype torch.float32 and shape [2435], and 'E' with the dtype torch.float32 and shape [775, 2435]. 

    \item Gemm Scaling Hardtanh GELU:the known variables are 'A' with the dtype torch.float16 and shape [919, 884], 'G' with the dtype torch.float16 and shape [], 'H' with the dtype torch.float16 and shape [1447, 884], 'I' with the dtype torch.float16 and shape [1447], and 'F' with the dtype torch.float16 and shape [919, 1447]. 

    \item Matmul Scaling ResidualAdd: the known variables are 'A' with the dtype torch.float32 and shape [457, 3520], 'C' with the dtype torch.float32 and shape [457, 2265], 'G' with the dtype torch.float32 and shape [2265], 'H' with the dtype torch.float32 and shape [2265], 'I' with the dtype torch.float32 and shape [2265, 3520], 'J' with the dtype torch.float32 and shape [2265], and 'F' with the dtype torch.float32 and shape [457, 2265]. 

    \item Conv3d Multiply InstanceNorm Clamp Multiply Max: the known variables are 'A' with the dtype torch.float32 and shape [89, 22, 50, 3, 28], 'C' with the dtype torch.float32 and shape [], 'D' with the dtype torch.float32 and shape [], 'M' with the dtype torch.float32 and shape [3, 1, 1, 1], 'N' with the dtype torch.float32 and shape [3, 1, 1, 1], 'O' with the dtype torch.float32 and shape [3, 22, 1, 1, 1], 'Q' with the dtype torch.float32 and shape [3], 'R' with the dtype torch.float32 and shape [3], and 'K' with the dtype torch.float32 and shape [89, 50, 3, 28]. 

    \item ELU: the known variables are 'A' with the dtype torch.float16 and shape [784, 29541], and 'C' with the dtype torch.float16 and shape [784, 29541]. 

   \item  Matmul Swish Sum GroupNorm:  the known variables are 'A' with the dtype torch.float64 and shape [422, 297], 'H' with the dtype torch.float64 and shape [64], 'I' with the dtype torch.float64 and shape [64], 'J' with the dtype torch.float64 and shape [64, 297], 'K' with the dtype torch.float64 and shape [64], 'M' with the dtype torch.float64 and shape [64], 'N' with the dtype torch.float64 and shape [64], and 'G' with the dtype torch.float64 and shape [422, 64]. 

    \item Conv2d InstanceNorm Divide: the known variables are 'A' with the dtype torch.float16 and shape [28, 40, 223, 198], 'F' with the dtype torch.float16 and shape [], 'G' with the dtype torch.float16 and shape [12, 40, 1, 1], 'H' with the dtype torch.float16 and shape [12], 'I' with the dtype torch.float16 and shape [12], and 'E' with the dtype torch.float16 and shape [28, 12, 223, 198]. 

    \item Conv2d Multiply LeakyReLU GELU:  the known variables are 'A' with the dtype torch.float32 and shape [87, 54, 446, 95], 'G' with the dtype torch.float32 and shape [3, 1, 1], 'H' with the dtype torch.float32 and shape [3, 1, 1], 'I' with the dtype torch.float32 and shape [3, 54, 1, 1], and 'F' with the dtype torch.float32 and shape [87, 3, 446, 95]. 

  \item   Matmul BatchNorm BiasAdd Divide Swish: the known variables are 'A' with the dtype torch.float64 and shape [800, 824], 'K' with the dtype torch.float64 and shape [1], 'M' with the dtype torch.float64 and shape [], 'N' with the dtype torch.float64 and shape [1], 'O' with the dtype torch.float64 and shape [926, 824], 'Q' with the dtype torch.float64 and shape [926], 'R' with the dtype torch.float64 and shape [926], 'S' with the dtype torch.float64 and shape [926], and 'J' with the dtype torch.float64 and shape [800, 926]. 

   \item  Matmul Sum Max AvgPool LogSumExp LogSumExp:  the known variables are 'A' with the dtype torch.float32 and shape [905, 3614], 'J' with the dtype torch.float32 and shape [1475, 3614], 'K' with the dtype torch.float32 and shape [1475], and 'I' with the dtype torch.float32 and shape [905, 1].

   \item  Matrix vector multiplication: the known variables are 'A' with the dtype torch.float64 and shape [784, 778], 'C' with the dtype torch.float64 and shape [778, 1], and 'D' with the dtype torch.float64 and shape [784, 1]. 

  \item   Conv2d Min Tanh Tanh:  the known variables are 'A' with the dtype torch.float64 and shape [216, 14, 40, 219], 'H' with the dtype torch.float64 and shape [30, 14, 2, 2], and 'G' with the dtype torch.float64 and shape [216, 1, 39, 218]. 

    \item Gemm GroupNorm Swish Multiply Swish: the known variables are 'A' with the dtype torch.float64 and shape [581, 1021], 'J' with the dtype torch.float64 and shape [128], 'K' with the dtype torch.float64 and shape [128], 'M' with the dtype torch.float64 and shape [128, 1021], 'N' with the dtype torch.float64 and shape [128], 'O' with the dtype torch.float64 and shape [128], 'Q' with the dtype torch.float64 and shape [128], and 'I' with the dtype torch.float64 and shape [581, 128]. 

   \item  softmax: the known variables are 'A' with the dtype torch.float64 and shape [604, 24802], and 'C' with the dtype torch.float64 and shape [604, 24802]. 

   \item  Gemm Scale BatchNorm:  the known variables are 'A' with the dtype torch.float16 and shape [515, 668], 'H' with the dtype torch.float16 and shape [363], 'I' with the dtype torch.float16 and shape [363], 'J' with the dtype torch.float16 and shape [363, 668], 'K' with the dtype torch.float16 and shape [363], 'M' with the dtype torch.float16 and shape [363], 'N' with the dtype torch.float16 and shape [363], and 'G' with the dtype torch.float16 and shape [515, 363]. 
   
 \item Standard Matrix Multiplication :the known variables are 'A' with the dtype torch.float16 and shape [268, 222], 'C' with the dtype torch.float16 and shape [222, 2364], and 'D' with the dtype torch.float16 and shape [268, 2364]. 
Standard matrix multiplication:  the known variables are 'A' with the dtype torch.float16 and shape [268, 222], 'C' with the dtype torch.float16 and shape [222, 2364], and 'D' with the dtype torch.float16 and shape [268, 2364]. 
    
    \item Gemm BatchNorm GELU GroupNorm Mean ReLU: the known variables are 'A' with the dtype torch.float32 and shape [307, 2520], 'K' with the dtype torch.float32 and shape [64, 2520], 'M' with the dtype torch.float32 and shape [64], 'N' with the dtype torch.float32 and shape [64], 'O' with the dtype torch.float32 and shape [64], 'Q' with the dtype torch.float32 and shape [64], 'R' with the dtype torch.float32 and shape [64], and 'J' with the dtype torch.float32 and shape [307, 1]. 

   \item  Matmul with irregular shapes:  the known variables are 'A' with the dtype torch.float32 and shape [484, 2508], 'C' with the dtype torch.float32 and shape [2508, 1224], and 'D' with the dtype torch.float32 and shape [484, 1224]. 

   \item  GQA mean:  the known variables are 'A' with the dtype torch.float64 and shape [77, 180, 112], 'Ak' with the dtype torch.bool and shape [1, 1, 180, 180], 'Am' with the dtype torch.float64 and shape [112, 112], 'An' with the dtype torch.float64 and shape [112], 'Ao' with the dtype torch.float64 and shape [56, 112], 'Aq' with the dtype torch.float64 and shape [56], 'Ar' with the dtype torch.float64 and shape [56, 112], 'As' with the dtype torch.float64 and shape [56], 'Aw' with the dtype torch.float64 and shape [112, 112], 'Ax' with the dtype torch.float64 and shape [112], and 'Aj' with the dtype torch.float64 and shape [77, 1, 112]. 

   \item  MHA mean: the known variables are 'A' with the dtype torch.float64 and shape [88, 189, 196], 'Ah' with the dtype torch.bool and shape [1, 1, 189, 189], 'Ai' with the dtype torch.float64 and shape [588, 196], 'Aj' with the dtype torch.float64 and shape [588], 'Ak' with the dtype torch.float64 and shape [196, 196], 'Am' with the dtype torch.float64 and shape [196], and 'Ag' with the dtype torch.float64 and shape [88, 1, 196]. 

   \item  Matmul with transposed B: the known variables are 'A' with the dtype torch.float16 and shape [1024, 512], 'C' with the dtype torch.float16 and shape [768, 512], and 'E' with the dtype torch.float16 and shape [1024, 768]. 

   \item  Conv2d AvgPool Sigmoid Sum: the known variables are 'A' with the dtype torch.float64 and shape [58, 33, 34, 29], 'G' with the dtype torch.float64 and shape [50, 33, 1, 1], and 'F' with the dtype torch.float64 and shape [58]. 

   \item  Gemm ReLU Divide:  the known variables are 'A' with the dtype torch.float64 and shape [63, 1160], 'F' with the dtype torch.float64 and shape [], 'G' with the dtype torch.float64 and shape [1990, 1160], 'H' with the dtype torch.float64 and shape [1990], and 'E' with the dtype torch.float64 and shape [63, 1990]. 

   \item  ReLU:  the known variables are 'A' with the dtype torch.float16 and shape [694, 17889], and 'C' with the dtype torch.float16 and shape [694, 17889]. 

   \item  Matmul Sigmoid Sum: the known variables are 'A' with the dtype torch.float32 and shape [106, 3856], 'F' with the dtype torch.float32 and shape [2265, 3856], 'G' with the dtype torch.float32 and shape [2265], and 'E' with the dtype torch.float32 and shape [106, 1]. 

    \item conv depthwise separable 2D:  the known variables are 'A' with the dtype torch.float16 and shape [16, 3, 256, 256], 'E' with the dtype torch.float16 and shape [3, 1, 3, 3], 'F' with the dtype torch.float16 and shape [64, 3, 1, 1], and 'D' with the dtype torch.float16 and shape [16, 64, 254, 254]. 

    \item Conv2d GELU GlobalAvgPool: the known variables are 'A' with the dtype torch.float16 and shape [96, 29, 241, 115], 'H' with the dtype torch.float16 and shape [20, 29, 1, 1], and 'G' with the dtype torch.float16 and shape [96, 20]. 

    \item GQA Gemm:  the known variables are 'A' with the dtype torch.float64 and shape [1006, 10, 288], 'Ak' with the dtype torch.bool and shape [1, 1, 10, 10], 'Am' with the dtype torch.float64 and shape [288, 288], 'An' with the dtype torch.float64 and shape [288], 'Ao' with the dtype torch.float64 and shape [18, 288], 'Aq' with the dtype torch.float64 and shape [18], 'Ar' with the dtype torch.float64 and shape [18, 288], 'As' with the dtype torch.float64 and shape [18], 'Aw' with the dtype torch.float64 and shape [288, 288], 'Ax' with the dtype torch.float64 and shape [288], 'Ay' with the dtype torch.float64 and shape [297, 288], and 'Aj' with the dtype torch.float64 and shape [1006, 10, 297]. 

   \item  Matmul AvgPool GELU Scale Max:  the known variables are 'A' with the dtype torch.float16 and shape [845, 2795], 'K' with the dtype torch.float16 and shape [], 'M' with the dtype torch.float16 and shape [1148, 2795], 'N' with the dtype torch.float16 and shape [1148], and 'J' with the dtype torch.float16 and shape [845]. 

    \item Softplus: the known variables are 'A' with the dtype torch.float32 and shape [966, 15929], and 'C' with the dtype torch.float32 and shape [966, 15929]. 

   \item  Max Pooling 3D:  the known variables are 'A' with the dtype torch.float16 and shape [16, 32, 64, 64, 64], and 'C' with the dtype torch.float16 and shape [16, 32, 30, 30, 30]. 

   \item  Matmul GELU Softmax:  the known variables are 'A' with the dtype torch.float32 and shape [34, 405], 'F' with the dtype torch.float32 and shape [3679, 405], 'G' with the dtype torch.float32 and shape [3679], and 'E' with the dtype torch.float32 and shape [34, 3679]. 

   \item  Matmul for upper triangular matrices:  the known variables are 'A' with the dtype torch.float16 and shape [894, 894], 'C' with the dtype torch.float16 and shape [894, 894], and 'G' with the dtype torch.float16 and shape [894, 894]. 

    \item Sigmoid: the known variables are 'A' with the dtype torch.float32 and shape [701, 5484], and 'C' with the dtype torch.float32 and shape [701, 5484]. 

    \item Max reduction over a dimension:  the known variables are 'A' with the dtype torch.float32 and shape [1182, 276, 82], and 'D' with the dtype torch.float32 and shape [276, 82]. 

    \item HuberLoss:  the known variables are 'A' with the dtype torch.float16 and shape [228, 115], 'C' with the dtype torch.float16 and shape [228, 115], and 'D' with the dtype torch.float16 and shape []. 
  
    \item SELU: the known variables are 'A' with the dtype torch.float32 and shape [264, 114], and 'C' with the dtype torch.float32 and shape [264, 114]. 

    \item Gemm Divide Sum Scaling: the known variables are 'A' with the dtype torch.float32 and shape [867, 1407], 'H' with the dtype torch.float32 and shape [229, 1407], 'I' with the dtype torch.float32 and shape [], and 'G' with the dtype torch.float32 and shape [867, 1]. 
  
    \item Gemm GroupNorm Hardtanh:  the known variables are 'A' with the dtype torch.float64 and shape [576, 1440], 'I' with the dtype torch.float64 and shape [8, 1440], 'J' with the dtype torch.float64 and shape [8], 'K' with the dtype torch.float64 and shape [8], 'M' with the dtype torch.float64 and shape [8], and 'H' with the dtype torch.float64 and shape [576, 8, 1, 2]. 

    \item Matmul GroupNorm LeakyReLU Sum: the known variables are 'A' with the dtype torch.float32 and shape [132, 2982], 'G' with the dtype torch.float32 and shape [1024, 2982], 'H' with the dtype torch.float32 and shape [1024], 'I' with the dtype torch.float32 and shape [1024], 'J' with the dtype torch.float32 and shape [1024], and 'F' with the dtype torch.float32 and shape [132, 1024]. 

    \item Matmul with diagonal mat:  the known variables are 'A' with the dtype torch.float16 and shape [377], 'C' with the dtype torch.float16 and shape [377, 291], and 'E' with the dtype torch.float16 and shape [377, 291]. 

    \item Matmul Scale ResidualAdd Clamp LogSumExp Mish: the known variables are 'A' with the dtype torch.float16 and shape [94, 2974], 'J' with the dtype torch.float16 and shape [], 'K' with the dtype torch.float16 and shape [862, 2974], 'M' with the dtype torch.float16 and shape [862], and 'I' with the dtype torch.float16 and shape [94, 1]. 
 
    \item Matmul Add Swish Tanh GELU Hardtanh: the known variables are 'A' with the dtype torch.float16 and shape [244, 2994], 'J' with the dtype torch.float16 and shape [38], 'K' with the dtype torch.float16 and shape [38], 'M' with the dtype torch.float16 and shape [38, 2994], 'N' with the dtype torch.float16 and shape [38], and 'I' with the dtype torch.float16 and shape [244, 38]. 

    \item GELU:  the known variables are 'A' with the dtype torch.float64 and shape [480, 15255], and 'C' with the dtype torch.float64 and shape [480, 15255]. 
    Min reduction over a dimension:  the known variables are 'A' with the dtype torch.float16 and shape [947, 817, 20], and 'D' with the dtype torch.float16 and shape [817, 20]. 

    \item Matmul for lower triangular matrices: the known variables are 'A' with the dtype torch.float32 and shape [935, 935], 'C' with the dtype torch.float32 and shape [935, 935], and 'G' with the dtype torch.float32 and shape [935, 935]. 

    \item L2Norm:  the known variables are 'A' with the dtype torch.float32 and shape [93, 12038], and 'D' with the dtype torch.float32 and shape [93, 12038]. 

    \item Conv3d Scaling Tanh Multiply Sigmoid:  the known variables are 'A' with the dtype torch.float16 and shape [61, 62, 9, 54, 3], 'H' with the dtype torch.float16 and shape [43, 1, 1, 1], 'I' with the dtype torch.float16 and shape [43, 1, 1, 1], 'J' with the dtype torch.float16 and shape [43, 1, 1, 1], 'K' with the dtype torch.float16 and shape [43, 1, 1, 1], 'M' with the dtype torch.float16 and shape [43, 62, 3, 3, 3], and 'G' with the dtype torch.float16 and shape [61, 43, 7, 52, 1]. 

    \item Gemm GroupNorm Min BiasAdd:  the known variables are 'A' with the dtype torch.float32 and shape [76, 1833], 'K' with the dtype torch.float32 and shape [1, 1, 1, 2], 'M' with the dtype torch.float32 and shape [1, 1, 1, 2], 'N' with the dtype torch.float32 and shape [1024, 1833], 'O' with the dtype torch.float32 and shape [1024], 'Q' with the dtype torch.float32 and shape [1024], 'R' with the dtype torch.float32 and shape [1024], and 'J' with the dtype torch.float32 and shape [76, 1, 1, 2]. 

    \item HardTanh:  the known variables are 'A' with the dtype torch.float16 and shape [603, 26865], and 'C' with the dtype torch.float16 and shape [603, 26865]. 

    \item conv3d conv3d padding: the known variables are 'A' with the dtype torch.float64 and shape [134, 4, 14, 2013, 9], 'C' with the dtype torch.float64 and shape [134, 4, 14, 2013, 9], 'G' with the dtype torch.float64 and shape [1, 4, 1, 2, 2], and 'F' with the dtype torch.float64 and shape [134, 1, 8, 2018, 9]. 

    \item Conv3d Mish Tanh: the known variables are 'A' with the dtype torch.float64 and shape [103, 40, 22, 8, 16], 'F' with the dtype torch.float64 and shape [32, 40, 2, 2, 2], and 'E' with the dtype torch.float64 and shape [103, 32, 8, 3, 6]. 
    
    \item Conv3d Min Softmax:  the known variables are 'A' with the dtype torch.float16 and shape [41, 5, 5, 3, 19], 'G' with the dtype torch.float16 and shape [36, 5, 3, 3, 3], and 'F' with the dtype torch.float16 and shape [41, 3, 1, 17].

    \item Conv2d BatchNorm Scaling: the known variables are 'A' with the dtype torch.float32 and shape [105, 55, 35, 144], 'H' with the dtype torch.float32 and shape [], 'I' with the dtype torch.float32 and shape [6, 55, 6, 6], 'J' with the dtype torch.float32 and shape [6], 'K' with the dtype torch.float32 and shape [6], and 'G' with the dtype torch.float32 and shape [105, 6, 30, 139]. 
    
\end{enumerate}

\appendix

\end{document}